\documentclass[10pt,twocolumn,letterpaper]{article}

\usepackage[pagenumbers]{cvpr} 

\usepackage{graphicx}
\usepackage{amsmath}
\usepackage{amssymb}
\usepackage{booktabs}
\usepackage{color, colortbl}
\usepackage{xcolor}
\usepackage{xspace}
\usepackage{graphbox}

\usepackage{amsmath,amsfonts,bm}

\def\eqref#1{equation~\ref{#1}}

\def\1{\bm{1}}

\def\vzero{{\bm{0}}}

\def\vtheta{{\bm{\theta}}}

\def\vdelta{{\bm{\delta}}}

\def\vw{{\bm{w}}}
\def\vx{{\bm{x}}}

\DeclareMathAlphabet{\mathsfit}{\encodingdefault}{\sfdefault}{m}{sl}
\SetMathAlphabet{\mathsfit}{bold}{\encodingdefault}{\sfdefault}{bx}{n}

\newcommand{\R}{\mathbb{R}}

\usepackage[pagebackref,breaklinks,colorlinks]{hyperref}

\usepackage[capitalize]{cleveref}
\crefname{section}{Sec.}{Secs.}
\Crefname{section}{Section}{Sections}
\Crefname{table}{Table}{Tables}
\crefname{table}{Tab.}{Tabs.}

\definecolor{LightCyan}{rgb}{0.88,1,1}
\definecolor{Gray}{rgb}{.8,.8,.8}
\definecolor{header}{gray}{1}
\definecolor{subheader2}{rgb}{1, 0.9, 0.8}
\definecolor{subheader}{rgb}{0.87, 1, 0.82}
\definecolor{lightorange}{rgb}{1., .95, .55}

\usepackage{glossaries}
\usepackage[inline]{enumitem}
\glsdisablehyper

\definecolor{TartOrange}{HTML}{ff2e35}
\definecolor{Orange}{HTML}{ff7825}
\definecolor{Mango}{HTML}{ffc013}
\definecolor{AppleGreen}{HTML}{7cb81b}
\definecolor{Blue}{HTML}{1173b0}
\definecolor{BdazzledBlue}{HTML}{2e58a5}
\definecolor{Purple}{HTML}{5b3590}
\definecolor{Sunglow}{HTML}{FFCA3A}
\definecolor{TableRow}{gray}{0.9}

\newcommand{\imagenet}{\textsc{ImageNet}\xspace}
\newcommand{\imageneta}{\textsc{ImageNet-A}\xspace}
\newcommand{\imagenetc}{\textsc{ImageNet-C}\xspace}
\newcommand{\imagenetr}{\textsc{ImageNet-R}\xspace}
\newcommand{\imagenetreal}{\textsc{ImageNet-Real}\xspace}
\newcommand{\imagenetvtwo}{\textsc{ImageNet-V2}\xspace}
\newcommand{\imagenetsketch}{\textsc{ImageNet-Sketch}\xspace}
\newcommand{\conflictstimuli}{\textsc{Conflict Stimuli}\xspace}

\newcommand{\cifar}{\textsc{Cifar-10}\xspace}

\newcommand{\resnet}{\textsc{ResNet}\xspace}
\newcommand{\wrn}{\textsc{WideResNet}\xspace}
\newcommand{\vit}{\textsc{ViT}\xspace}
\newcommand{\vitb}{\textsc{ViT-B16}\xspace}

\newcommand{\Max}{\textsc{Max}\xspace}
\newcommand{\sat}{\textsc{Sat}\xspace}
\newcommand{\apgd}{\textsc{AutoPGD}\xspace}

\newcommand{\norm}[1]{\lVert#1\rVert}

\newlength{\newl}

\newcolumntype{C}[1]{>{\centering\arraybackslash}p{#1}}
\newcolumntype{L}[1]{>{\raggedright\arraybackslash}p{#1}}
\newcolumntype{R}[1]{>{\raggedleft\arraybackslash}p{#1}}

\newcommand{\squishlist}{
   \begin{list}{$\bullet$}
    {\setlength{\itemsep}{0pt} \setlength{\parsep}{0pt}
     \setlength{\topsep}{0pt} \setlength{\partopsep}{0pt}
     \setlength{\leftmargin}{1em} \setlength{\labelwidth}{1em}
     \setlength{\labelsep}{0.5em}}}
\newcommand{\squishend}{
    \end{list}}

\begin{document}

\title{
Seasoning Model Soups for Robustness \\ to Adversarial and Natural Distribution Shifts
}

\author{
Francesco Croce\textsuperscript{*} \\ University of T{\"u}bingen
\and
Sylvestre-Alvise Rebuffi \\ DeepMind
\and
Evan Shelhamer \\ DeepMind
\and
Sven Gowal \\ DeepMind
}
\maketitle

\def\thefootnote{*}\footnotetext{Work done during an internship at DeepMind.}\def\thefootnote{\arabic{footnote}}

\begin{abstract}
Adversarial training is widely used to make classifiers robust to a specific threat or adversary, such as $\ell_p$-norm bounded perturbations of a given $p$-norm.
However, existing methods for training classifiers robust to multiple threats require knowledge of all attacks during training and remain vulnerable to unseen distribution shifts.
In this work, we describe how to obtain adversarially-robust \emph{model soups} (i.e., linear combinations of parameters) that smoothly trade-off robustness to different $\ell_p$-norm bounded adversaries.
We demonstrate that such soups allow us to control the type and level of robustness, and can achieve robustness to all threats without jointly training on all of them. 
In some cases, the resulting model soups are more robust to a given $\ell_p$-norm adversary than the constituent model specialized against that same adversary.
Finally, we show that adversarially-robust model soups can 
be a viable tool to adapt to distribution shifts from a few examples.
\end{abstract}

\section{Introduction}

Deep networks have achieved great success on several computer vision tasks and have even reached super-human accuracy~\cite{krizhevsky_imagenet_2012,he_deep_2015}.
However, the outputs of such models are often brittle, and tend to perform poorly on inputs that differ from the distribution of inputs at training time, in a condition known as \emph{distribution shift}~\cite{quinonero2009dataset}.
Adversarial perturbations are a prominent example of this condition: small, even imperceptible, changes to images can alter predictions to cause errors~\cite{szegedy_intriguing_2013, biggio_evasion_2013}.
In addition to adversarial inputs, it has been noted that even natural shifts, e.g. different weather conditions, can significantly reduce the accuracy of even the best vision models \cite{hendrycks_benchmarking_2018,geirhos_generalisation_2018,recht_imagenet_2019}.
Such drops in accuracy are undesirable for robust deployment, and so a lot of effort has been invested in correcting them.
Adversarial training \cite{madry_towards_2017} and its extensions \cite{zhang_theoretically_2019, gowal_improving_2021, rebuffi_data_2021} are currently the most effective methods to improve empirical robustness to adversarial attacks.
Similarly, data augmentation is the basis of several techniques that improve robustness to non-adversarial/natural shifts \cite{hendrycks_augmix_2019, calian_defending_2021, erichson_noisymix_2022}.
While significant progress has been made on defending against a specific, selected type of perturbations (whether adversarial or natural), it is still challenging to make a single model robust to a broad set of threats and shifts.
For example, a classifier adversarially-trained for robustness to $\ell_p$-norm bounded attacks is still vulnerable to attacks in other $\ell_q$-threat models \cite{tramer_adversarial_2019,kang_testing_2019}.
Moreover, methods for simultaneous robustness to multiple attacks require jointly training on all \cite{maini_adversarial_2019, madaan2021learning} or a subset of them \cite{croce2022adversarial}. Most importantly, controlling the trade-off between different types of robustness (and nominal performance) remains difficult and requires training several classifiers.

Inspired by \emph{model soups}~\cite{wortsman_model_2022}, which interpolate the parameters of a set of vision models to achieve state-of-the-art accuracy on \imagenet, we investigate the effects of 
interpolating robust image classifiers.
We complement their original recipe for soups by own study of how to pre-train, fine-tune, and combine the parameters of models adversarially-trained against $\ell_p$-norm bounded attacks for different $p$-norms.
To create models for soups, we pre-train a single robust model and fine-tune it to the target threat models (using the efficient technique of~\cite{croce2022adversarial}).
We then establish that it is possible to smoothly trade-off robustness to different threat models by moving in the convex hull 
of the parameters of each robust classifier, while achieving competitive performance with methods that train on multiple $p$-norm adversaries simultaneously.
Unlike alternatives, our soups can uniquely \textit{(1)} choose the level of robustness to each threat model without any further training and \textit{(2)} quickly adapt 
to new unseen attacks or shifts by simply tuning the weighting of the soup.
 
Previous works \cite{xie_adversarial_2019, kireev_effectiveness_2021, herrmann_pyramid_2021} have shown that adversarial training with $\ell_p$-norm bounded attacks can help to improve performance on natural shifts if carefully tuned.
We show that model soups of diverse classifiers, with different types of robustness, offer greater flexibility for finding models that perform well across various shifts, such as \imagenet variants.
Furthermore, we show that a limited number of images of the new distribution are sufficient to select the weights of such a model soup.
Examining the composition of the best soups brings insights about which features are important for each dataset and shift.
Finally, while the capability of selecting a model specific to each image distribution is a main point of our model soups, we also show that it is possible to jointly select a soup for average performance across several \imagenet variants to achieve better accuracy than adversarial and self-supervised baselines~\cite{herrmann_pyramid_2021, he_masked_2021}.

\noindent\textbf{Contributions.} In summary, we show that soups
\squishlist
\item can merge nominal and $\ell_p$-robust models (for various $p$): efficient fine-tuning from one robust model obtains a set of models with diverse robustness~\cite{croce2022adversarial} and compatible parameters for creating model soups ~\cite{wortsman_model_2022} 
(Sec.~\ref{sec:how_to_make_soups}),
\item can control the level of robustness to each threat model and achieve, without more training, competitive performance against multi-norm robustness training
(Sec.~\ref{sec:soups_for_lp_robustness}),
\item are not limited to interpolation, but can find more effective classifiers by extrapolation (Sec.~\ref{sec:improving_individual_threat_models}),
\item enable adaptation to unseen distribution shifts on only a few examples (Sec.~\ref{sec:soups_for_distribution_shifts}).
\squishend

\section{Related Work}

\textbf{Adversarial robustness to multiple threat models.}
Most methods focus on achieving robustness in a single threat model, i.e. to a specific type of attacks used during training.
However, this is not sufficient to obtain robustness to unseen attacks.
As a result, further work aims to train classifiers for simultaneous robustness to multiple attacks, and the most popular scenario considers a set of $\ell_p$-norm bounded perturbations.
The most successful methods~\cite{tramer_adversarial_2019, maini_adversarial_2019, madaan2021learning} are based on adversarial training and differ in how the multiple threats are combined. 
Notably, all need to use every attack at training time. 
To reduce the computational cost of obtaining multiply-robust models, one can fine-tune a singly-robust model by one of the above mentioned methods~\cite{croce2022adversarial}, even for only a small number of epochs. 
Our soups are more flexible by skipping simultaneous adversarial training across multiple attacks.

\textbf{Adversarial training for distribution shifts.}
While robustness to adversarial attacks and natural shifts are not the same goal, previous works nevertheless show that it is possible to leverage adversarial training with $\ell_p$-norm bounded attacks to improve performance on the common corruptions of \cite{hendrycks_benchmarking_2018}.
First, AdvProp \cite{xie_adversarial_2019} co-trains models on clean and adversarial images (in the $\ell_\infty$-threat model) with dual normalization layers that specialize to each type of input.
The clean branch of the dual model achieves higher accuracy than nominal training on \imagenet and its variants.
Similar results are obtained by Pyramid-AT \cite{herrmann_pyramid_2021} by its design of a specific attack to adversarially train vision transformers.
Finally, \cite{kireev_effectiveness_2021} carefully selecting the size of the adversarial perturbations, i.e. their $\ell_\infty$- or $\ell_2$-norm, for standard adversarial training \cite{madry_towards_2017} achieves competitive performance on common corruptions on \cifar and \imagenet-100.

\textbf{Model soups.} 
Ensembling or averaging the parameters of intermediate models found during training is an effective technique to improve both clean accuracy \cite{huang2017snapshot,izmailov_averaging_2018} and robustness \cite{gowal_uncovering_2020, rebuffi_data_2021}.
Recently, \cite{wortsman_model_2022} propose \emph{model soups} which interpolate the parameters of networks fine-tuned with different hyperparameters configurations from the same pre-trained model.
This yields improved classification accuracy on \imagenet.
Along the same line, \cite{ilharco2022patching} fine-tune a model trained on \imagenet on several new image classification datasets, and show that interpolating the original and fine-tuned parameters yields classifiers that perform well on all tasks.

\setlength{\newl}{.52\columnwidth}
\begin{figure*}[t]\centering\small
\begin{tabular}{ccc}
\textcolor{blue}{soup:} $\vtheta_\infty + \vtheta_{\infty \rightarrow 2}$, \textcolor{orange}{soup:} $\vtheta_2 + \vtheta_{2 \rightarrow \infty}$ & \textcolor{blue}{soup:} $\vtheta_\infty + \vtheta_{\infty \rightarrow 1}$, \textcolor{orange}{soup:} $\vtheta_1 + \vtheta_{1 \rightarrow \infty}$ & \textcolor{blue}{soup:} $\vtheta_2 + \vtheta_{2 \rightarrow 1}$, \textcolor{orange}{soup:} $\vtheta_1 + \vtheta_{1 \rightarrow 2}$\\
\includegraphics[width=\newl]{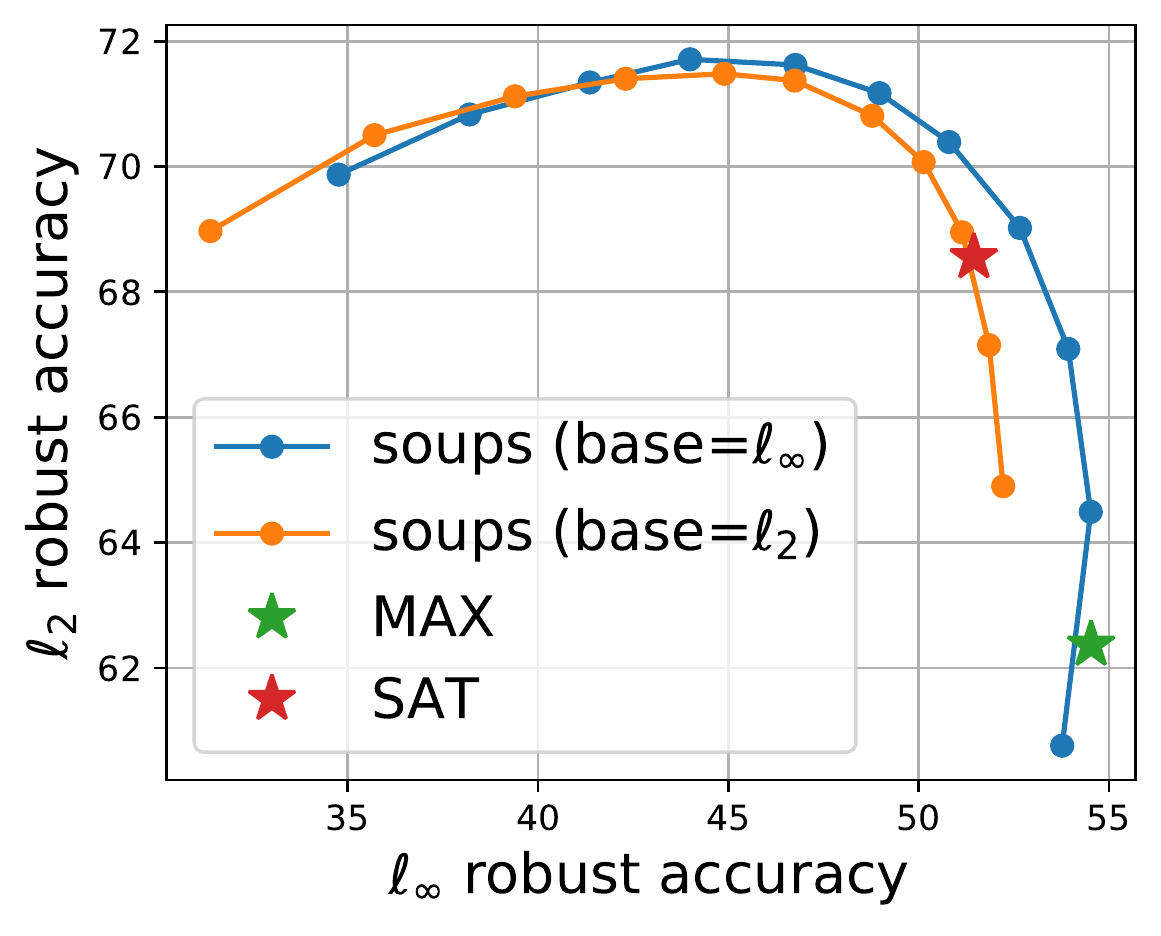} &
\includegraphics[width=\newl]{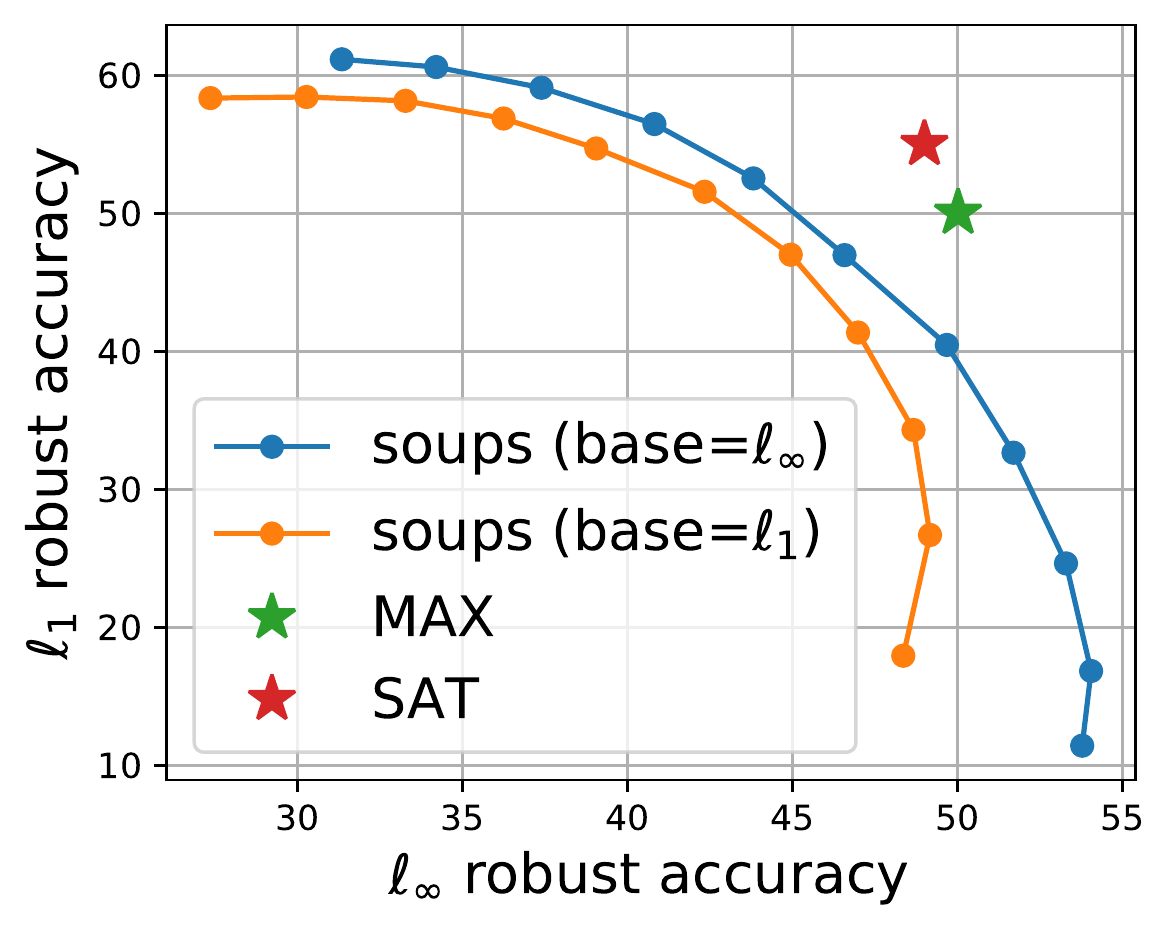} &
\includegraphics[width=\newl]{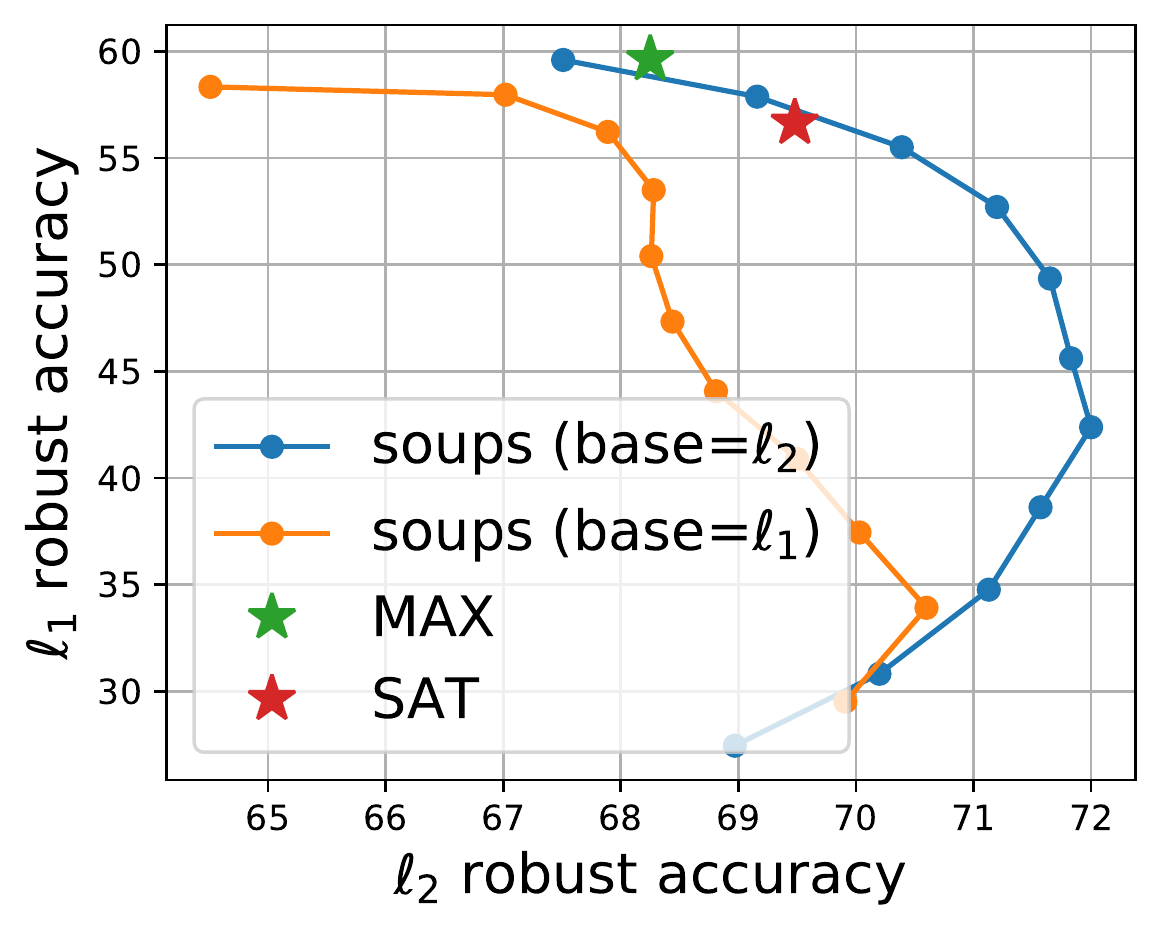}
\end{tabular}
\caption{
\textbf{Soups of two models on \cifar:} 
for all pairs $(p, q)$ we show the $\ell_p$- vs $\ell_q$-robust accuracy of the soups $w\cdot\vtheta_p + (1 - w) \cdot \vtheta_{p\rightarrow q}$ and $w\cdot\vtheta_q + (1 - w) \cdot \vtheta_{q\rightarrow p}$ varying $w \in [0, 1]$.
We also show results for \Max and \sat with simultaneous use of the two threat models.} \label{fig:soups_twomodels_cifar10}
\end{figure*}

\begin{figure*}\centering\small
\tabcolsep=1.3pt
\begin{tabular}{cccc}
soup: $\vtheta_\infty + \vtheta_{\infty \rightarrow 2}$ & soup: $\vtheta_\infty + \vtheta_{\infty \rightarrow 1}$ & soup: $\vtheta_{\infty \rightarrow 2} + \vtheta_{\infty \rightarrow 1}$ & soup: $\vtheta_\infty + \vtheta_{\infty \rightarrow \textrm{nominal}}$ \\
\includegraphics[width=\newl]{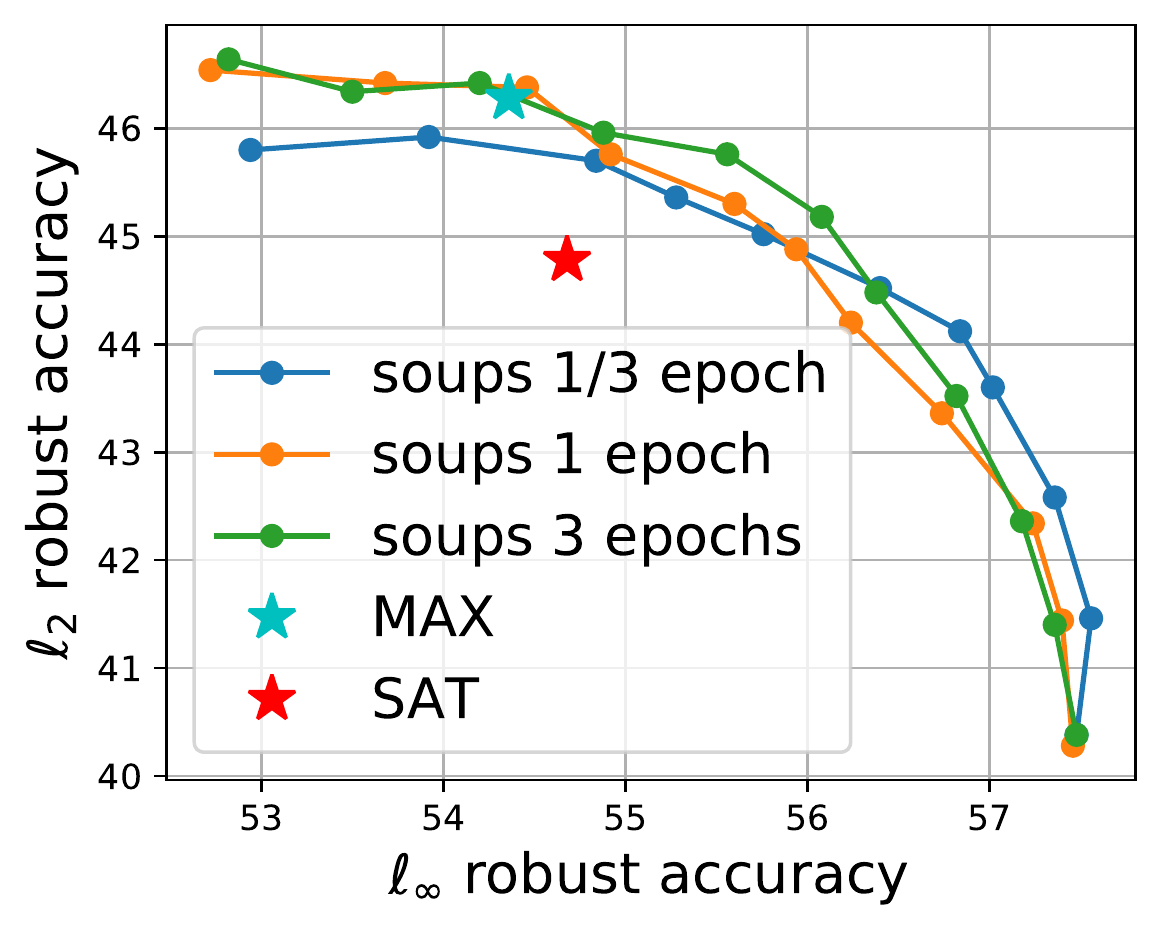} &
\includegraphics[width=\newl]{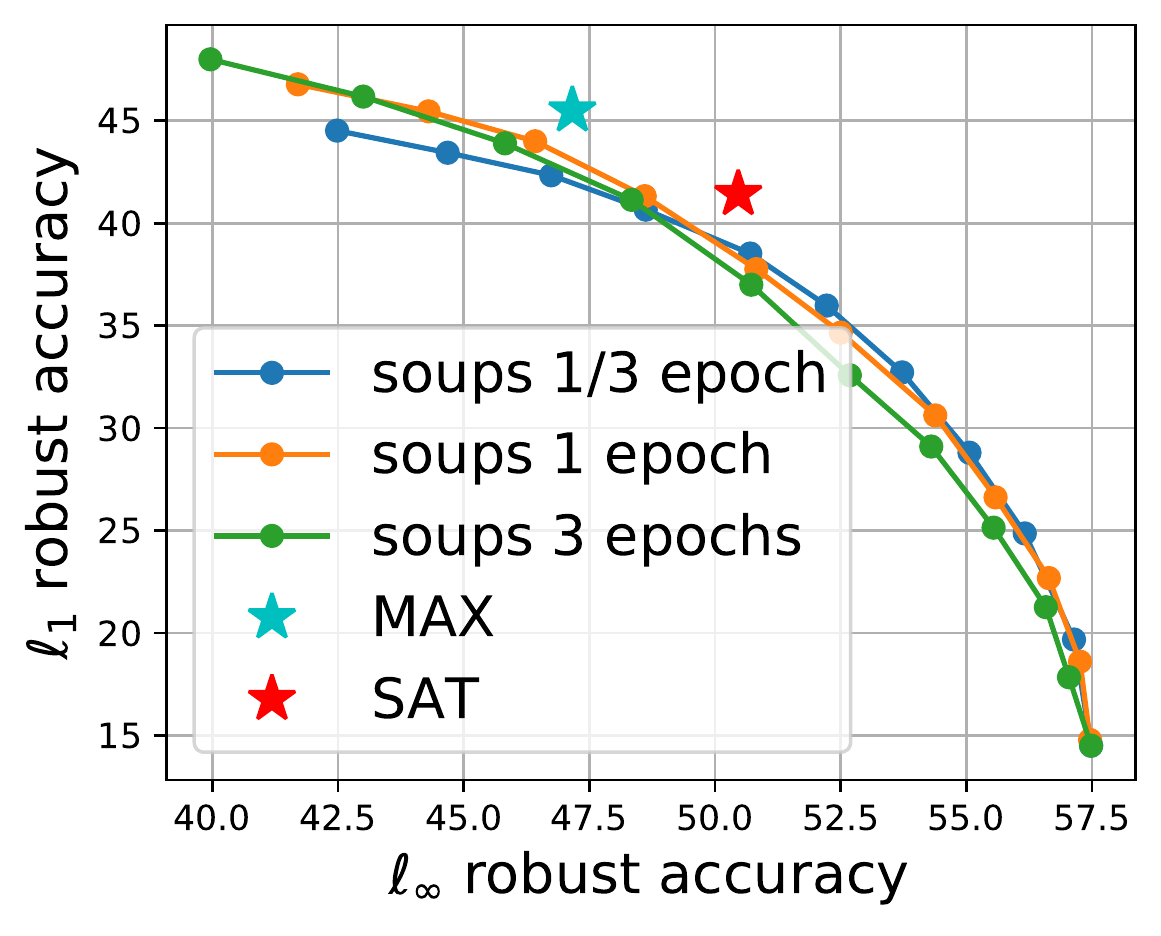} &
\includegraphics[width=\newl]{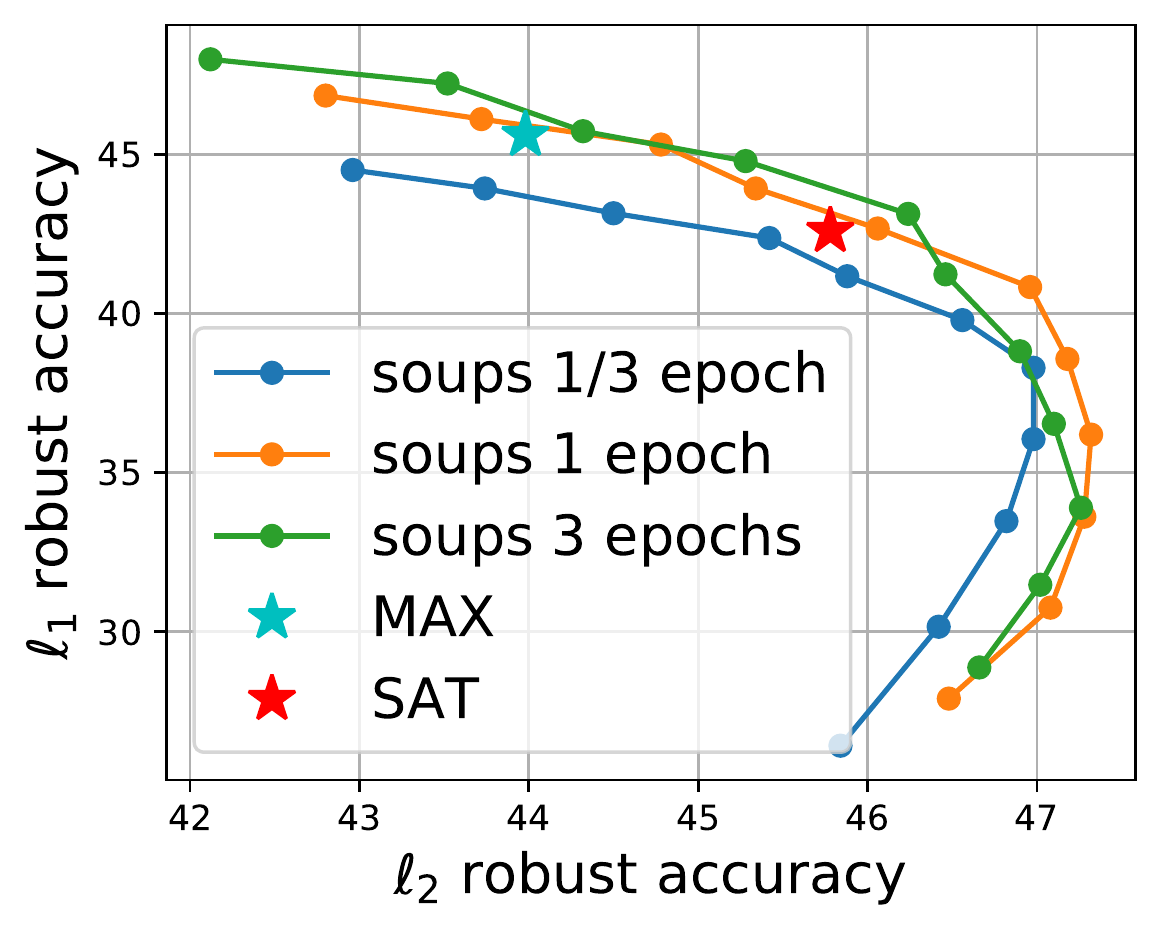} & \includegraphics[width=\newl]{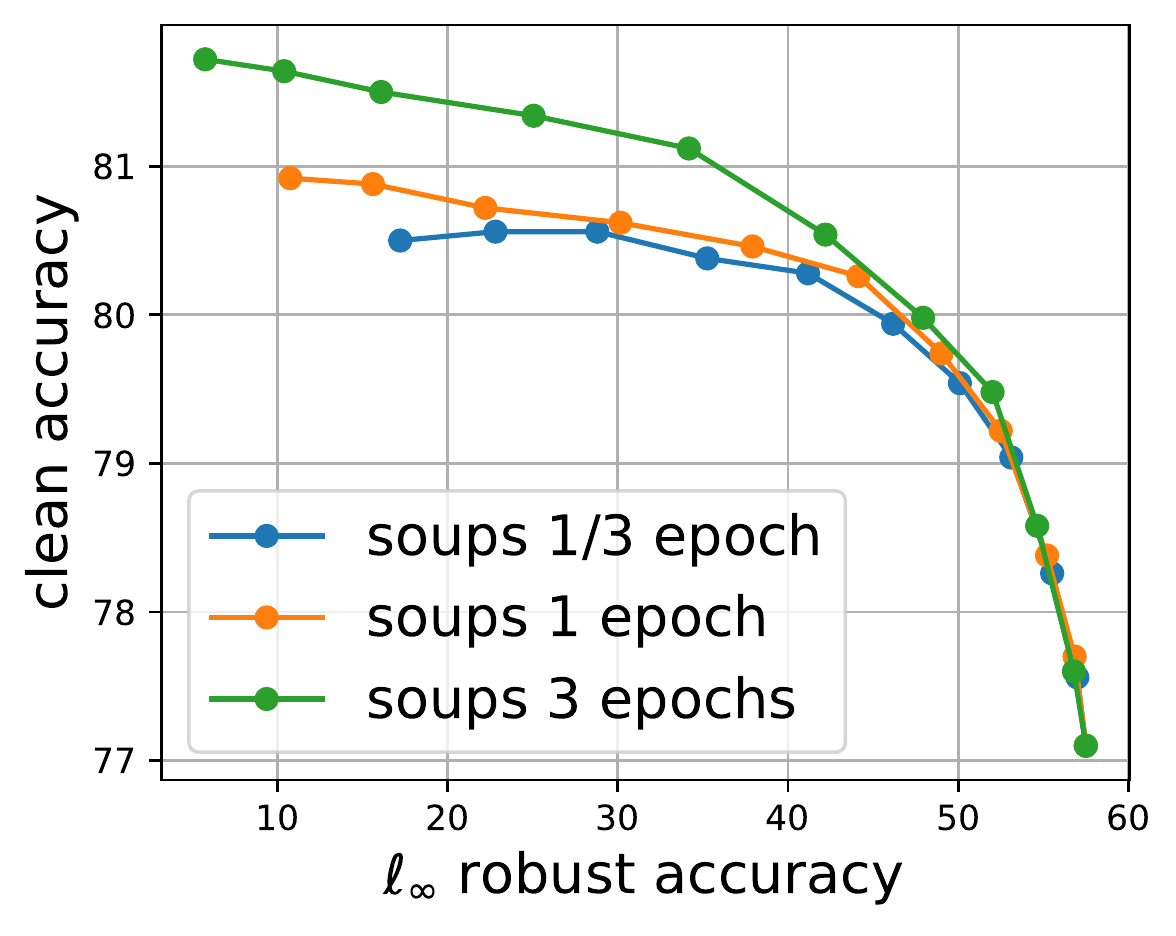}
\end{tabular}
\caption{\textbf{Soups of two models on \imagenet:} 
for $p \in \{2, 1\}$ we show the $\ell_p$- vs $\ell_\infty$-robust accuracy of the soups $w\cdot\vtheta_\infty + (1 - w) \cdot \vtheta_{\infty \rightarrow p}$ varying $w \in [0, 1]$ (first and second columns). Moreover, we show the soups obtained combining $\vtheta_{\infty \rightarrow 2}$ and $\vtheta_{\infty \rightarrow 1}$ (third), or $\vtheta_\infty$ and $\vtheta_{\infty \rightarrow \text{nominal}}$ (fourth). For the case of two $\ell_p$-threat models, we also show the results of fine-tuning models with \Max and \sat.} \label{fig:soups_two_models_vits}
\end{figure*}

\section{Model Interpolation across Different Tasks}
\label{sec:how_to_make_soups}
In the following, we formally introduce the two main components of our procedure to merge adversarially robust models: \textit{(1)} obtaining models which can interpolated by fine-tuning a single $\ell_p$-robust classifiers, and \textit{(2)} interpolation of their weights to balance their different types of robustness.
We highlight that our setup diverges from that of prior works about parameters averaging: in fact, both \cite{wortsman_model_2022, ilharco2022patching} combine models fine-tuned on the same task, i.e. achieving high classification accuracy of unperturbed images, either on a fixed dataset and different hyperparameter configurations \cite{wortsman_model_2022}, or varying datasets \cite{ilharco2022patching}. 
In our case, the individual models are trained for robustness to different types of attacks, i.e. with distinct loss functions.

\subsection{Adversarial training and fine-tuning}
Let us denote $\mathcal{D} = \{(\vx_i, y_i)\}_i$ the training set, with  $\vx_i\in \R^d$ indicating an image and $y_i\in 
\{1, \ldots, K\}$ the corresponding label, and $\Delta: \R^d \rightarrow \R^d$ the function which characterizes a threat model, that maps an input $\vx$ to a set $\Delta(\vx)\subset \R^d$ of possible perturbed versions of the original image. For example, $\ell_p$-norm bounded adversarial attacks with budget $\epsilon > 0$ in the image space can be described by
\begin{align} \Delta(\vx) = \{\vdelta \in \R^d : \norm{\vdelta}_p \leq \epsilon, \vx + \vdelta \in[0, 1]^d\}. \label{eq:threat_models}
\end{align} Then, one can train a classifier $f: \vtheta \times \R^d \rightarrow \R^K$  parameterized by $\vtheta \in \Theta$ with adversarial training \cite{madry_towards_2017} by solving
\begin{align} \min_{\vtheta \in \Theta}\, \sum_{(\vx, y) \in \mathcal{D}}\,\max_{\vdelta \in \Delta(\vx)}\, \mathcal{L}(f(\vtheta, x + \vdelta), y),
\end{align}
for a given loss function $\mathcal{L}:\R^K \times \R^K \rightarrow \R$ (e.g., cross-entropy), with the goal of obtaining a model robust to the perturbations described by $\Delta$.  Note that this boils down to nominal training when $\Delta(\cdot) = \{\vzero\}$ and no perturbation is applied on the training images. We are interested in the case where multiple threat models are available, and indicate with $\Delta_\text{nominal}$ nominal training, and $\Delta_p$ for $p\in\{\infty, 2, 1\}$ the perturbations with bounded $\ell_p$-norm as described in Eq.~\ref{eq:threat_models}. We focus on such tasks since they are the most common choices for adversarial defenses, in particular by methods focusing on multiple norm robustness \cite{tramer_adversarial_2019, maini_adversarial_2019}.

Notably, it is possible to efficiently obtain a model robust to $\Delta_q$ by fine-tuning for a single (or few) epoch with adversarial training w.r.t. $\Delta_q$ a classifier pre-trained to be robust in $\Delta_p$, with $p\neq q$ and $p, q\in\{\infty, 2, 1\}$~\cite{croce2022adversarial}.
For example, in this way one can efficiently derive specialized classifiers $\Delta_2, \Delta_1, \Delta_\text{nominal}$ for each task from a single model $f$ robust w.r.t. $\ell_\infty$. 
However, this does not work well when a nominal classifier is used as starting point for the short fine-tuning.
In the following, we denote with $\vtheta_{p\rightarrow q}$ the parameters resulting from fine-tuning to $\Delta_q$ a base model $\vtheta_p$.

\subsection{Merging different types of robustness via linear combinations in parameter space}

We want to explore the properties of the models obtained by taking linear combinations of the parameters of classifiers with different types of robustness. To do so, there needs to be a correspondence among the parameters of different individual networks: \cite{wortsman_model_2022} achieve this by merging differently fine-tuned versions of the same pre-trained model.
As mentioned above, fine-tuning an $\ell_p$-robust classifier allows to change it to achieve robustness in a new threat model \cite{croce2022adversarial}: we exploit such property to create \emph{model soups}, as named by \cite{wortsman_model_2022}.
Formally, we create a model soup from $n$ individual networks with parameters $\vtheta^1, \dots, \vtheta^n$ with weights $\vw = (w_1, \ldots, w_n) \in \R^n$ as
\begin{align} \vtheta^\vw = \sum_{i=1}^n\, w_i \cdot \vtheta^i,
\end{align}
and the corresponding classifier is given by $f(\vtheta^\vw, \cdot):\R^d\rightarrow \R^K$.
While any choice of $\vw$ is possible, we focus on the case of affine combinations, i.e. $\sum_i w_i=1$.
Moreover, we consider soups which are either convex combinations of the individual models, with $w_1, \dots, w_n \geq 0$, or obtained by extrapolations, i.e. with negative elements in $\vw$.

\section{Soups for $\ell_p$-robustness}\label{sec:soups_for_lp_robustness}
We measure adversarial robustness in the $\ell_p$-threat model with bounds $\epsilon_p$: on \cifar we use $\epsilon_\infty=8/255$, $\epsilon_2=128/255$, $\epsilon_1=12$, on \imagenet $\epsilon_\infty=4/255$, $\epsilon_2=4$, $\epsilon_1=255$.
If not specified otherwise, we use the full test set for \cifar and 5000 images from the \imagenet validation set, and attack by \apgd \cite{croce_reliable_2020, croce_mind_2021} with 40 steps.
More details are provided in App.~\ref{sec:app_exp_details}.

\subsection{Soups with two threat models}

\textbf{\cifar.} 
We explore the effect of interpolating two models robust to different $\ell_p$-norm bounded attacks for $p\in\{\infty, 2, 1\}$. 
We consider classifiers with \wrn-28-10 \cite{zagoruyko2016wide} architecture trained on \cifar. 
For every threat model, we first train a robust classifier with adversarial training from random initialization. 
Then, we fine-tune the resulting model with adversarial training on each of the other threat models for 10 epochs. 
In Fig.~\ref{fig:soups_twomodels_cifar10} we show, for each pair of threat models $(\Delta_p, \Delta_q)$, the trade-off of robust accuracy w.r.t. $\ell_p$ and $\ell_q$ for the soups
\[w \cdot \vtheta_p + (1 - w) \cdot \vtheta_{p\rightarrow q} \quad \text{for}\quad w \in [0, 1]\]
and symmetrically
\[w \cdot \vtheta_q + (1 - w) \cdot \vtheta_{q\rightarrow p} \quad \text{for}\quad w \in [0, 1].\]
Interpolating the parameters of models trained with a single $\ell_p$-norm 
controls the balance between the two types of robustness: for example, Fig.~\ref{fig:soups_twomodels_cifar10} (middle plot) shows that moving from $\vtheta_\infty$ to $\vtheta_{\infty \rightarrow 1}$ (blue curve), i.e. decreasing $w$ from 1 to 0 in the corresponding soup, progressively reduces the robust accuracy w.r.t. $\ell_\infty$ to improve robustness w.r.t. $\ell_1$.
Moreover, for similar threat models (i.e. the pairs $(\ell_2, \ell_1)$ and $(\ell_2, \ell_\infty)$) some intermediate networks are more robust than the extremes trained specifically for each threat model.

\textbf{\imagenet.} 
We now fine-tune 
a \vitb \cite{dosovitskiy_image_2020} robust w.r.t. $\ell_\infty$ on \imagenet 
to the other threat models, including nominal training, for either 1/3, 1 or 3 epochs.
Fig.~\ref{fig:soups_two_models_vits} shows that interpolation of parameters is effective even in this setup, and allows to easily balance nominal and robust accuracy (fourth plot).
Moreover, it is possible to create soups with two fine-tuned models, i.e. $\vtheta_{\infty \rightarrow 2}$ and $\vtheta_{\infty \rightarrow 1}$.
Finally, increasing the number of fine-tuning steps yields better performance in the target threat model, which in turn generally leads to better soups.

\textbf{Comparison to multi-norm robustness methods.} Fig.~\ref{fig:soups_twomodels_cifar10} and Fig.~\ref{fig:soups_two_models_vits} compare the performance of the model soups to that of models trained with methods for robustness in the union of multiple threat models.
In particular, we show the results of \Max \cite{tramer_adversarial_2019}, which computes perturbations for each threat model and trains on that attaining the highest loss, and \sat \cite{madaan2021learning}, which samples uniformly at random for each training batch the attack to use.
We train models with both methods for all pairs of threat models: as expected, \Max tends to focus on the most challenging threat model, sacrificing some robustness in the other one compared to \sat, since it uses only the strongest attack for each training point.
When training for $\ell_\infty$ and $\ell_1$, i.e. the extreme $\ell_p$-norms we consider, \Max and \sat models lie above the front drawn by the model soups, hinting that the more diverse the attacks, the more difficult it is to preserve high robustness to both.
In the other cases, 
both methods behave similarly to the soups.
The main advantage given by interpolation is however the option of moving along the front without additional training cost: while one might tune the trade-off between the robustness in the two threat models in \sat, e.g. changing the sampling probability, this would still require training a new classifier for each setup.

\begin{figure*}\centering\small
\begin{tabular}{c}
\rowcolor{Gray}
soup: $\vtheta_\infty + \vtheta_{\infty \rightarrow 2} + \vtheta_{\infty \rightarrow 1}$

\\
\includegraphics[width=2\columnwidth]{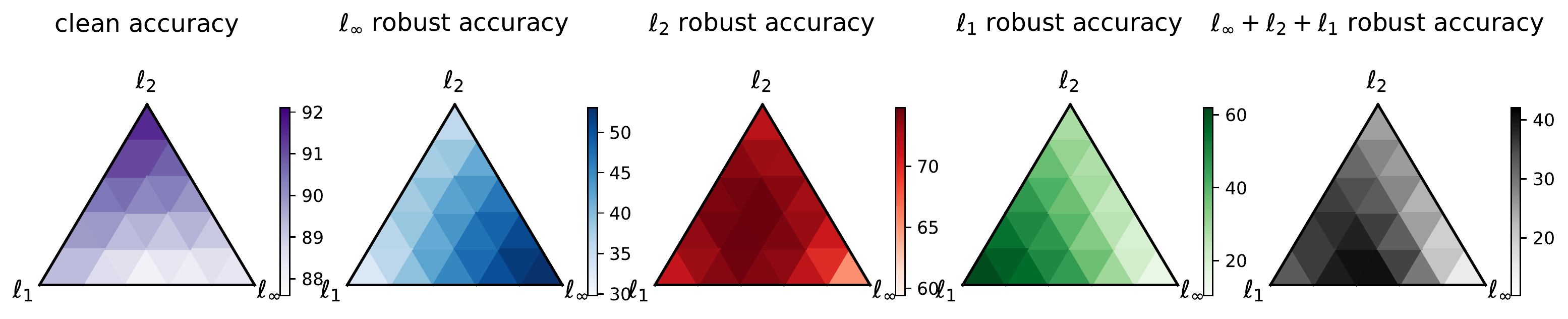}
\end{tabular}
\caption{
\textbf{Soups of three models on \cifar:} we fine-tune the model robust w.r.t. $\ell_\infty$ 
(with \wrn-28-10 architecture) to the other threat models for 10 epochs, and show clean accuracy (first column) and robust accuracy w.r.t. every threat model (second to fourth columns) and their union (last column) of the soups obtained as convex combinations of the three bases.} \label{fig:soups_threemodels_cifar10}
\end{figure*}

\begin{figure*}\centering\small
\begin{tabular}{c}
\rowcolor{Gray}
\resnet-50, 
soup: $\vtheta_\infty + \vtheta_{\infty \rightarrow 2} + \vtheta_{\infty \rightarrow 1}$\\ 
\includegraphics[width=2\columnwidth]{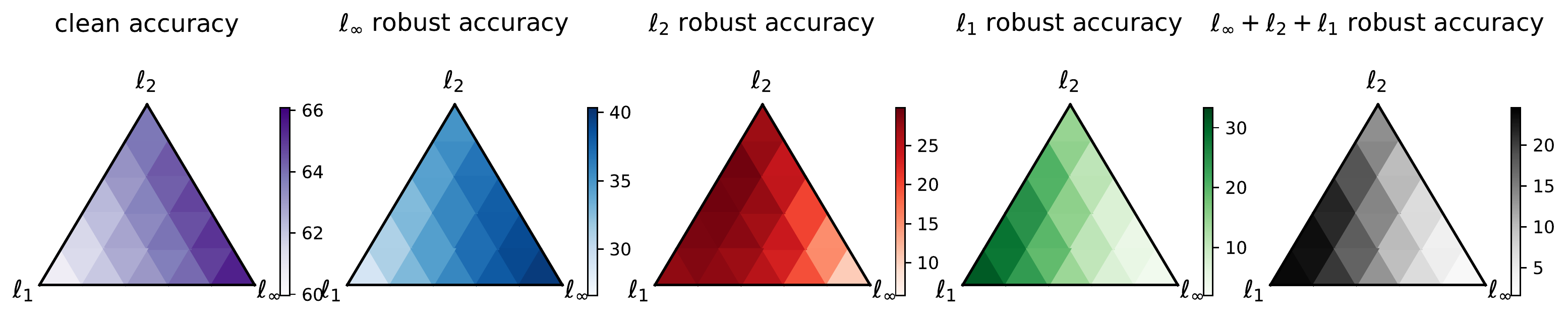}\\
\rowcolor{Gray}
\vitb, 
soup: $\vtheta_\infty + \vtheta_{\infty \rightarrow 2} + \vtheta_{\infty \rightarrow 1}$\\ 
\includegraphics[width=2\columnwidth]{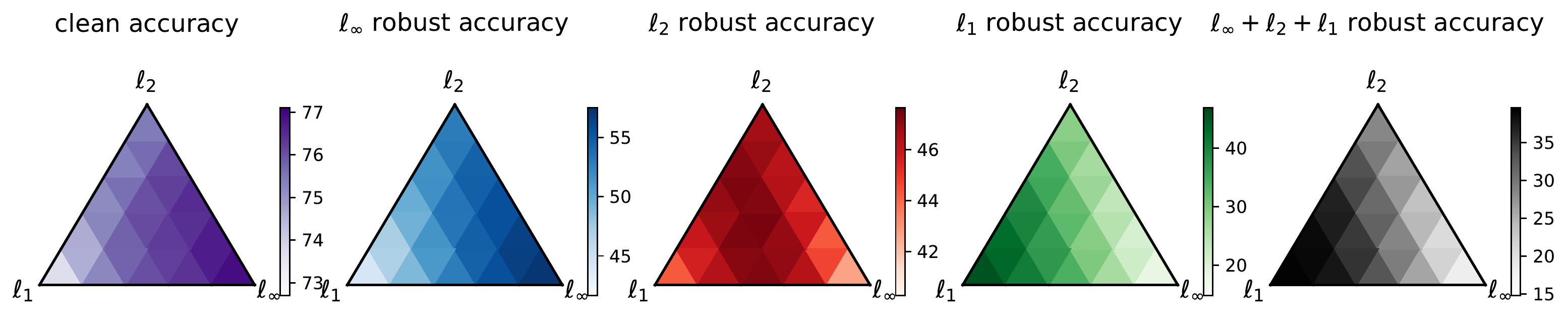}
\end{tabular}
\caption{
\textbf{Soups of three models on \imagenet:} we fine-tune the classifiers, with \resnet-50 (top row) and \vitb (bottom) as architecture, robust w.r.t. $\ell_\infty$ for 1 epoch to the other threat models, and show clean accuracy (first column) and robust accuracy in every threat model (second to fourth columns) and their union (last column) of the classifiers obtained as convex combinations of the three bases.} \label{fig:soups_threemodels_imagenet}
\end{figure*}

\subsection{Soups with three threat models}

We here study the convex combination of three models with different types of robustness. For \cifar we create soups with each $\ell_p$-robust classifier for $p\in\{\infty, 2, 1\}$ and its fine-tuned version into the other two threat models. We use the same models of the previous section, and sweep the interpolation weights $\vw\in\R^3$ such that $w_i\in\{0, 0.2, 0.4, 0.6, 0.8, 1\}$ and $\sum_i w_i=1$. In Fig.~\ref{fig:soups_threemodels_cifar10} and Fig.~\ref{fig:app_soups_threemodels_cifar10} (in the Appendix) we show clean accuracy (first column) and robust accuracy in $\ell_\infty$, $\ell_2$ and $\ell_1$ (second to fourth columns) and their union, when a point is considered robust only if it is such against all attacks (last column).

One can observe that, independently from the type of robustness of the base model (used as initialization for the fine-tuning), moving in the convex hull of the three parameters (e.g. $\vtheta_\infty + \vtheta_{\infty \rightarrow 2} + \vtheta_{\infty \rightarrow 1}$ in Fig.~\ref{fig:soups_threemodels_cifar10}) allows to smoothly control the trade-off of the different types of robustness.
Interestingly, the highest $\ell_2$-robustness is 
attained by intermediate soups, not by the model specifically fine-tuned w.r.t. $\ell_2$, suggesting that model soups might even be beneficial to robustness in individual threat models (see more below).
Moreover, the highest robustness in the union is given by interpolating only the models robust w.r.t. $\ell_\infty$ and $\ell_1$, which is in line with the observation of \cite{croce2022adversarial} that training for the extreme norms is sufficient for robustness in the union of the three threat models.
Although the robust accuracy in the union is lower than that of training simultaneously with all attacks e.g. with \Max (42.0\% vs 47.2\%), the model soups deliver competitive results without the need of co-training.

Finally, Fig.~\ref{fig:soups_threemodels_imagenet} shows that similar observations hold on \imagenet, where we create soups fine-tuning classifiers robust w.r.t. $\ell_\infty$, with either \resnet-50 \cite{he_deep_2015} or \vitb 
as architecture, for 1 epoch.

\subsection{Soups for improving individual threat model robustness}
\label{sec:improving_individual_threat_models}

\begin{figure}[h] \centering\small \tabcolsep=2pt
\begin{tabular}{*{2}{C{.5\columnwidth}}}
\cifar: $\vtheta_\infty + \vtheta_{\infty \rightarrow 2}$ & \imagenet: $\vtheta_\infty + \vtheta_{\infty \rightarrow 2}$ \\
\multicolumn{2}{c}{\includegraphics[width=1\columnwidth]{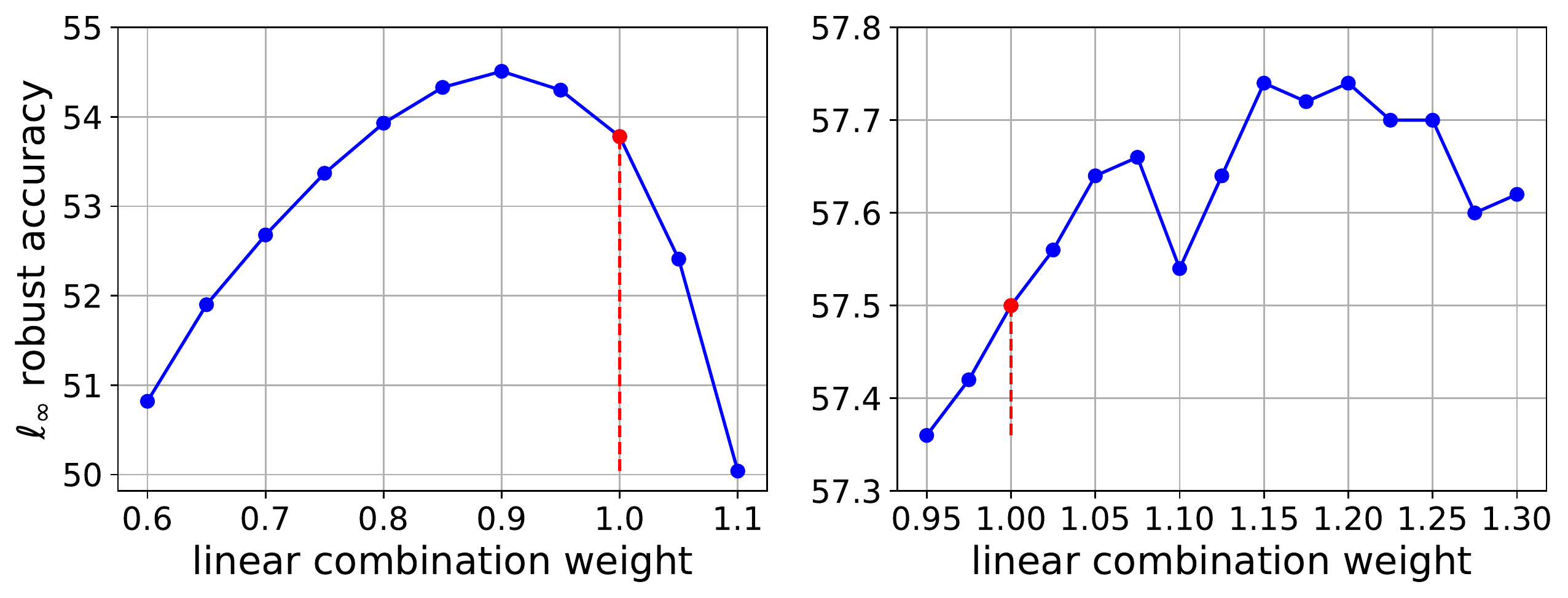}}
\end{tabular}
\caption{
\textbf{Improvement on single threat models:} we show the robust accuracy w.r.t. $\ell_\infty$ for the soups $w\cdot \vtheta_\infty + (1 - w)\cdot \vtheta_{\infty \rightarrow 2}$ for varied $w$.
The original model $\vtheta_\infty$ is highlighted in red.
}
\label{fig:single_threat_model}
\end{figure}

\begin{figure*}[t!]\centering\small
\includegraphics[width=2.1\columnwidth]{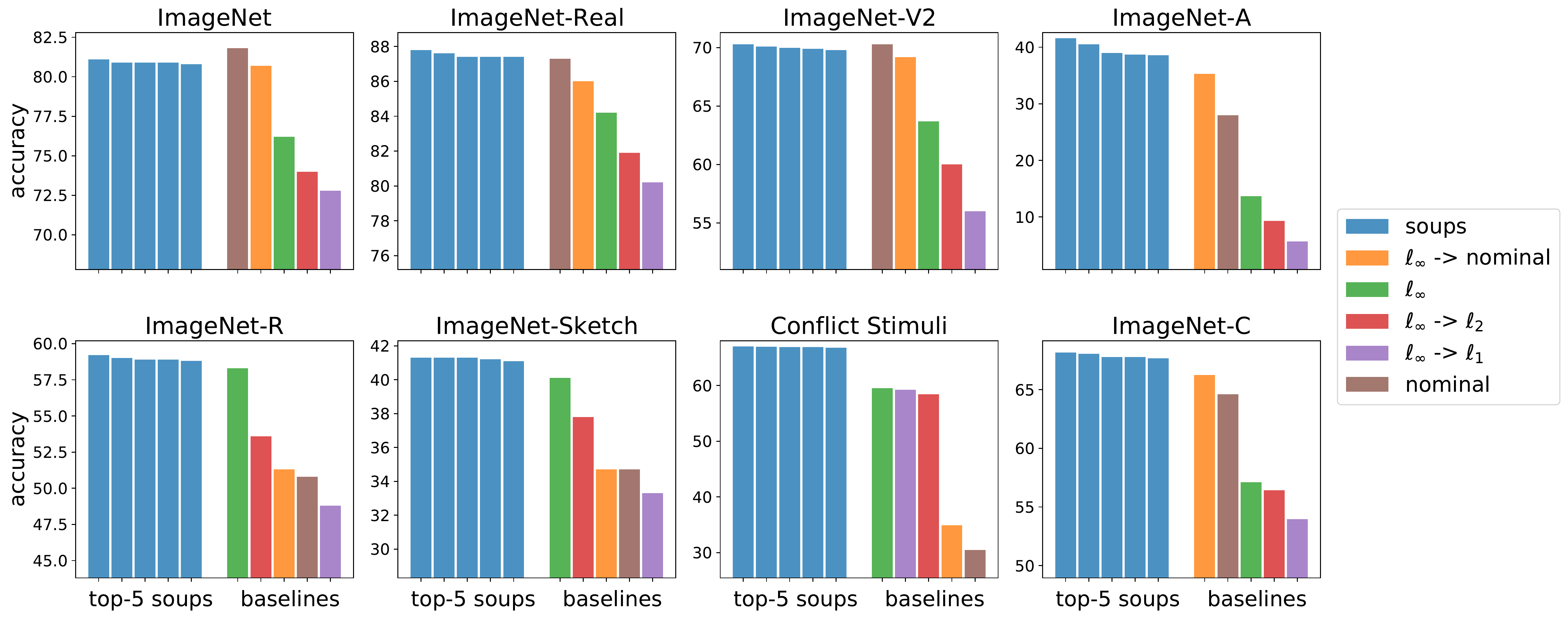}
\caption{\textbf{Soups on \imagenet variants:} for each dataset we plot the accuracy of the 5 best performing soups of the four base models  $\vtheta_\infty$, $\vtheta_{\infty \rightarrow 2}$, $\vtheta_{\infty \rightarrow 1}$ and $\vtheta_{\infty \rightarrow \text{nominal}}$, and of the individual classifiers. Additionally, we show the performance of an independently trained nominal model. All models are evaluated on the 1000 points used for the grid search of the best soups.} \label{fig:best_soups_vs_individualmodels}
\end{figure*}

We notice in Fig.~\ref{fig:soups_twomodels_cifar10} and Fig.~\ref{fig:soups_two_models_vits} that in a few cases the intermediate models obtain via parameters interpolation have higher robustness than the extreme ones, which are trained or fine-tuned with a specific threat model. As such, we analyze in more details the soups $w\cdot \vtheta_\infty + (1 - w)\cdot \vtheta_{\infty \rightarrow 2}$, i.e. using the original classifier robust in $\ell_\infty$ and the one fine-tuned to $\ell_2$, on both \cifar and \imagenet. Fig.~\ref{fig:single_threat_model} shows the robust accuracy w.r.t. $\ell_\infty$ when varying the value of $w$: in both case the original model $\vtheta_\infty$, highlighted in red, does not attain the best robustness. Interestingly, on \cifar the best soup is found with $w=0.9$, while for \imagenet 
with $w > 1$: this suggests that the model soups should not be constrained to the convex hull of the base models, and extrapolation can lead to improvement.

\section{Soups for Distribution Shifts} \label{sec:soups_for_distribution_shifts}
Prior works \cite{kireev_effectiveness_2021} have shown that adversarial training w.r.t. an $\ell_p$-norm is able to provide some improvement in the performance in presence of non-adversarial distribution shifts, e.g. the common corruptions of \imagenetc \cite{hendrycks_benchmarking_2018}.
However, to see such gains it is necessary to carefully select the threat model, for example which $\ell_p$-norm and size $\epsilon$ to bound the perturbations, to use during training.
The experiments in Sec.~\ref{sec:soups_for_lp_robustness} suggest that model soups of nominal and adversarially robust classifiers yield models with a variety of intermediate behaviors, and extrapolation might even deliver models which do not merely trade-off the robustness of the initial classifiers but amplify it.
This flexibility could suit adaptation to various distribution shifts: that is, the various corruption types might more closely resemble the geometry of different $\ell_p$-balls or their union.
Moreover, including a nominally fine-tuned model in the soup allows it to maintain, if necessary, high accuracy on the original dataset, which is often degraded by adversarial training~\cite{madry_towards_2017} or test-time adaptation on shifted data~\cite{niu2022efficient}.

\begin{table*}
\centering \footnotesize \tabcolsep=2pt
\begin{tabular}{l|c|*{8}{C{13mm}}|C{13mm}}
    
    \cellcolor{header} \textsc{Setup} & \cellcolor{header} \textsc{\# FP} & \cellcolor{header} \textsc{ImageNet}  & \cellcolor{header} \textsc{IN-Real} & \cellcolor{header} \textsc{IN-V2} & \cellcolor{header} \textsc{IN-A} & \cellcolor{header} \textsc{IN-R} & \cellcolor{header} \textsc{IN-Sketch} & \cellcolor{header} \textsc{Conflict Stimuli} & \cellcolor{header} \textsc{IN-C} & \cellcolor{header} \textsc{Mean}   \\ \toprule

    \multicolumn{11}{l}{\textbf{Baselines}} \\
    
    \toprule
    Nominal training & $\times 1$ & 82.64\% & 87.33\% & 71.42\% & 28.03\% &   47.94\% & 34.43\% & 30.47\% & 64.45\% & 55.84\%    \\
    Adversarial training & $\times 1$  & 76.88\% & 83.91\% & 64.81\% & 12.35\% & 55.76\% & 40.11\% & 59.45\% & 55.44\% & 56.09\%   \\
    Fine-tuned MAE-B16 & $\times 1$  & 83.10\% & 88.02\% & 72.80\% & 37.92\% & 49.30\% & 35.69\% & 27.81\% & 63.23\% & 57.23\%   \\
    AdvProp & $\times 1$ &83.39\% & 88.06\% & 73.17\% & 34.81\% & 53.04\% & 39.25\% & 38.98\% & 70.39\% & 60.14\% \\
    Pyramid-AT 
    & $\times 1$  & 83.14\% & 87.82\% & 72.53\% & 32.72\% & 51.78\% & 38.60\% & 37.27\% & 67.01\% & 58.86\%   \\
    Indep. networks ensemble & $\times 2$  & 82.86\% & 87.78\% & 71.73\% & 25.99\% & 54.20\% & 37.33\% & 46.41\% & 65.61\% & 58.99\%   \\

    Individual networks ensemble & $\times 4$ & 81.31\% & 86.97\% & 70.21\% & 23.13\% & 54.82\% & 39.51\% & 56.02\% & 68.17\% & 60.02\%\\
    \midrule
    
    \multicolumn{11}{l}{\textbf{Fixed grid search on 1000 images}} \\ \toprule
    Single soup  & $\times 1$ & 82.49\% & 87.85\% & 71.99\% & 34.31\% & 53.84\% & 39.84\% & 38.52\% & 66.82\% & 59.46\%\\
    \rowcolor{Gray}
    Dataset-specific soups & $\times 1$ & 82.29\% & 87.89\% & 71.95\% & 38.27\% & 56.39\% &40.73\% &67.03\% & 69.34\% & (64.24\%) \\
    \bottomrule

    \end{tabular}
    
    \caption{\textbf{Comparison on \imagenet variants:} we report the classification accuracy of various models on the \imagenet variants, together with the number of forward passes they require. The soups are selected via a fixed grid search on the interpolation weights with 1000 points for each dataset. The last row shows the results of the best found soup for each dataset.} \label{tab:imagenet_variants}
\end{table*}

\begin{figure*}\centering\small
\includegraphics[width=2.1\columnwidth]{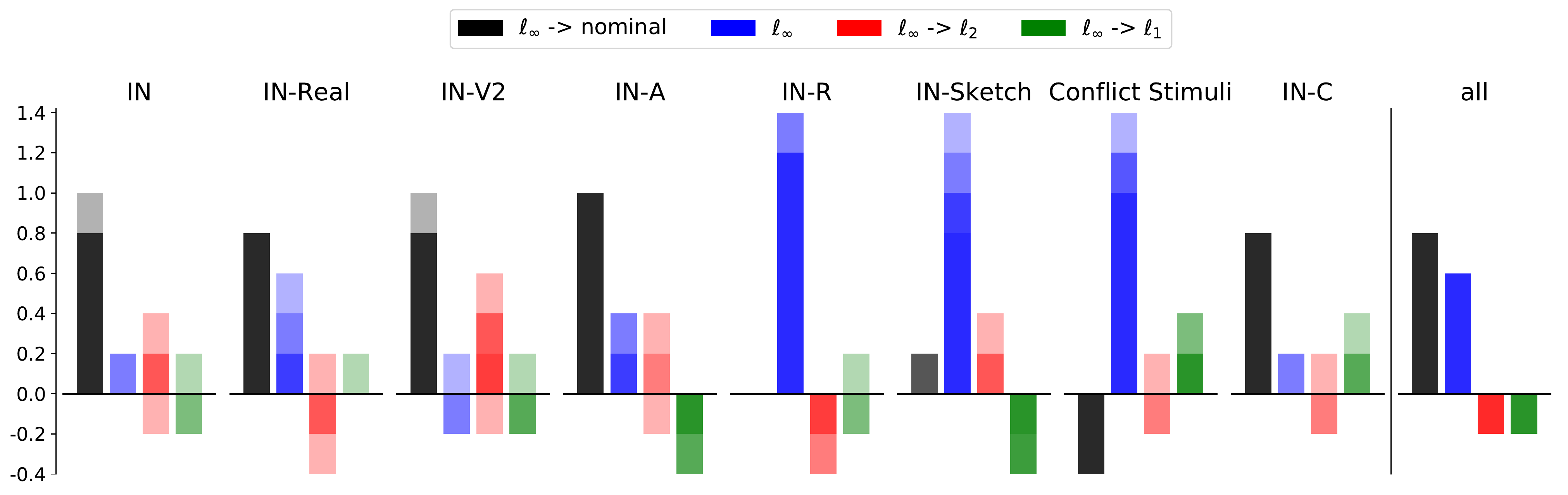}
\caption{
\textbf{Soup compositions on \imagenet variants:} for each dataset we plot the composition of the 5 best soups, i.e. the linear weights for the individual models, as measured by grid search over 1000 points on weights in the range $[-0.4, 1.4]$ with granularity $0.2$.  
Additionally we show the composition of the model achieving the best average accuracy over all 8 variants.} \label{fig:soups_composition}
\end{figure*}

\subsection{Soups for \imagenet variants}

\textbf{Setup.}
In the following, we use models soups consisting of robust \vit fine-tuned from $\ell_\infty$ to the other threat models, and one more \vit nominally fine-tuned for 100 epochs to obtain slightly higher accuracy on clean data.
For shifts, we consider several variants of \imagenet, providing a broad and diverse benchmark for our soups: \imagenetreal \cite{beyer_are_2020}, \imagenetvtwo \cite{recht_imagenet_2019}, \imagenetc \cite{hendrycks_benchmarking_2018},
\imageneta \cite{hendrycks_natural_2019}, \imagenetr \cite{hendrycks_many_2020}, \imagenetsketch \cite{wang2019learning}, and \conflictstimuli \cite{geirhos_imagenet-trained_2018}.
We consider the setting of few-shot supervised adaptation, with a small set of labelled images from each shift, which we use to select the best soups.

\textbf{Soup selection via grid search.}
Since evaluating the accuracy of many soups on the entirety of the datasets would be extremely expensive, we search for the best combination of the four models on a random subset of 1000 points from each dataset, with the exception of \conflictstimuli for which all 1280 images are used (for \imagenetc we use all corruption types and severities, then aggregate the results). 
Restricting our search to a subset also serves our aim of finding a model soup which generalizes to the new distribution by only seeing a few examples. 
We evaluate all the possible affine combinations 
with weights in the range $[-0.4, 1.4]$ with granularity $0.2$, which amounts to 460 models in total.
In Fig.~\ref{fig:best_soups_vs_individualmodels} we compare, for each dataset, the accuracy of the 5 best soups 
to that of each individual classifier used for creating the soups and of a nominal model trained independently: for all datasets apart from \imagenet the top soups outperform the individual models.
Moreover, we notice that the best individual model varies across datasets, indicating that it might be helpful to merge networks with different types of robustness.

\textbf{Comparison to existing methods.} Having selected the best soup for each variant (dataset-specific soups) on its chosen few-shot adaptation set, we evaluate the soup on the test set of the variant (results in Table~\ref{tab:imagenet_variants}).
We also evaluate the model soup that attains the best average case accuracy over the adaptation sets for all variants (single soup), in order to gauge the best performance of a single, general model soup.
We compare the soups to a nominal model, the $\ell_\infty$-robust classifier used in the soups, their ensemble, the Masked AutoEncoders of \cite{he_masked_2021}, AdvProp \cite{xie_adversarial_2019}, Pyramid-AT \cite{herrmann_pyramid_2021}, and the ensemble obtained by averaging the output (after softmax) of the four models included in the soups. 
Selecting the best soup on 1000 images of each datasets (results in the last row of Table~\ref{tab:imagenet_variants}) leads in 4 out of the 8 datasets to the best accuracy, and only slightly suboptimal values in the other cases: in particular, parameters interpolations is very effective on stronger shifts like \imagenetr and \conflictstimuli, where it attains almost 8\% better performance than the closest baseline.
Unsurprisingly, it is more challenging to improve on datasets like \imagenetvtwo which are very close to the original \imagenet.
Overall, The soup selected for best average accuracy (across all datasets) outperforms all baselines, except for the ensemble of four models (with $4\times$ the inference cost), and AdvProp, which requires co-training of clean and adversarial points. 
These results show that soups with robust classifiers are a promising avenue for quickly adapting to distribution shifts.

\begin{figure}[t]\centering
\includegraphics[width=1\columnwidth]{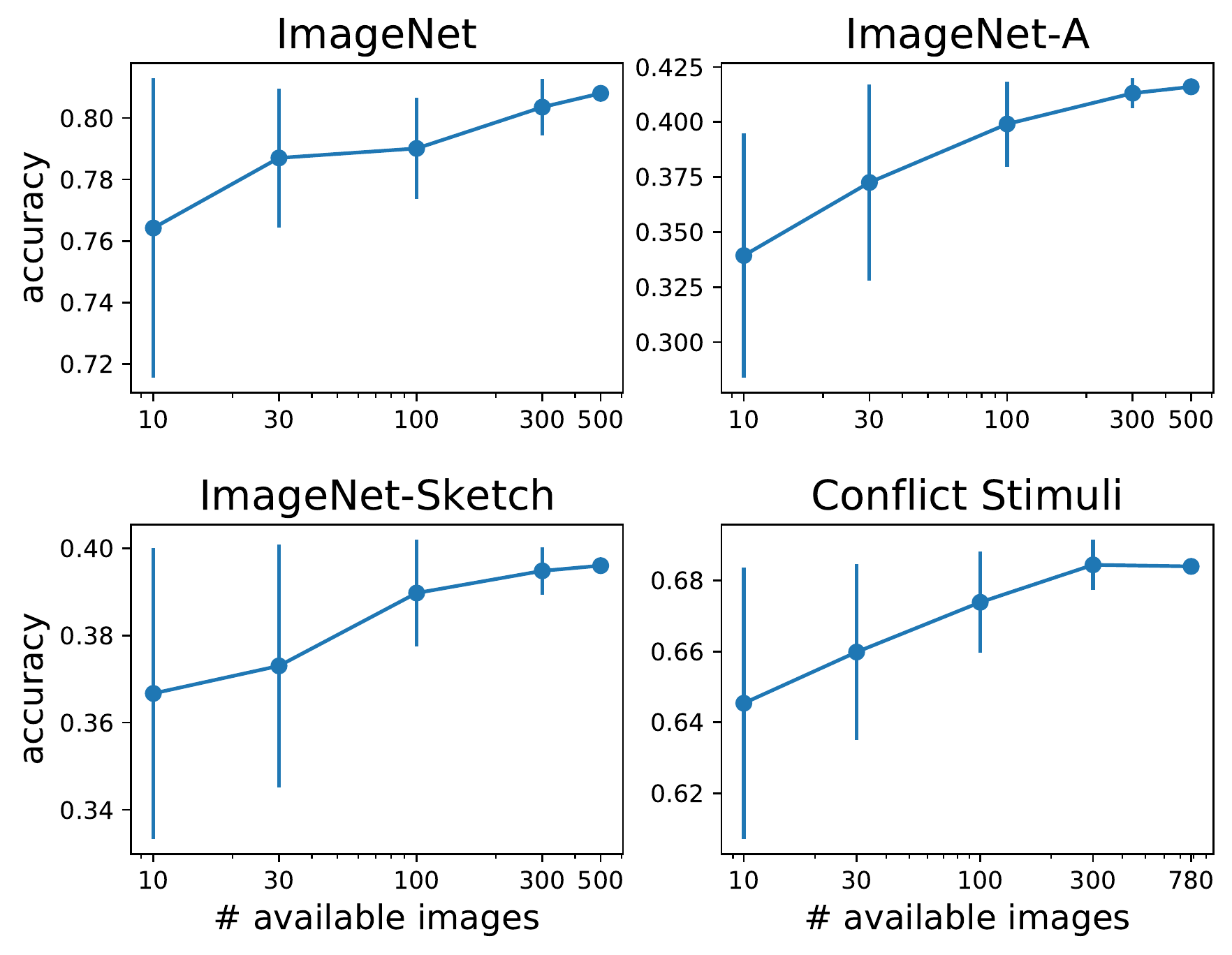}
\caption{
\textbf{Soup selection data efficiency:} we vary the number of images for selecting the best performing soups on different datasets.
For each case, we plot the average accuracy on a held-out test set, and its standard deviation over 50 trials.
} \label{fig:soup_selection_varying_images}
\end{figure}

\textbf{Composition of the soups.} To analyze which types of classifiers are most relevant for performance on every distribution shift, we plot in Fig.~\ref{fig:soups_composition} the breakdown of the weights of the five best soups 
(more intense colors indicate that the corresponding weight or a larger one is used more often in the top-5 soups).
First, one can see that the nominally fine-tuned model (in black) is dominant, with weights of 0.8 or 1, on \imagenet, \imagenetreal, \imagenetvtwo, \imageneta and \imagenetc: this could be expected since these datasets are closer to \imagenet itself, i.e. the distribution shift is smaller, which is what nominal training optimizes for (in fact, the nominal models achieve higher accuracy than adversarially trained ones on these datasets in Fig.~\ref{fig:best_soups_vs_individualmodels}). However, in all cases there is a contribution of some of the $\ell_p$-robust networks. On \imagenetr, \imagenetsketch and \conflictstimuli, the model robust w.r.t. $\ell_\infty$ plays instead the most relevant role, again in line with the results in Table~\ref{tab:imagenet_variants}. Interestingly, in the case of \conflictstimuli, the nominal classifier has a weight -0.4 (the smallest in the grid search) for all top performing soups: we hypothesize that this has the effect of reduce the texture bias typical of nominal model and emphasize the attention to shapes already important in adversarially trained classifiers. Finally, we show the composition of the soup which has the best average accuracy over all datasets (last column of Fig.~\ref{fig:soups_composition}), where the nominal and $\ell_\infty$-robust models have similar positive weight.

\textbf{How many images does it take to find a good soup?} 
To identify the practical limit of supervision for soup selection, we study the effect of varying the number of labelled images used to select the best soup on a new dataset.
For this analysis we randomly choose 500 images from the adaptation set used for the grid search to create a held-out test set.
From the remaining images, we uniformly sample $k$ elements, select the soup which performs best on such $k$ points, and then evaluate it on this  test set.
We repeat this procedure for $k\in\{10, 30, 100, 300, 500\}$ for 50 times each with different random seeds.
In Fig.~\ref{fig:soup_selection_varying_images} we plot the average accuracy on the held-out test set, with standard deviation, when varying $k$: increasing the number of points above 100 achieves high test accuracy with limited variance.
This suggests that soup selection, and thus model adaptation, can be carried out with as few as 100 examples of the new distribution.

\begin{figure}[t]\centering
\includegraphics[width=1\columnwidth]{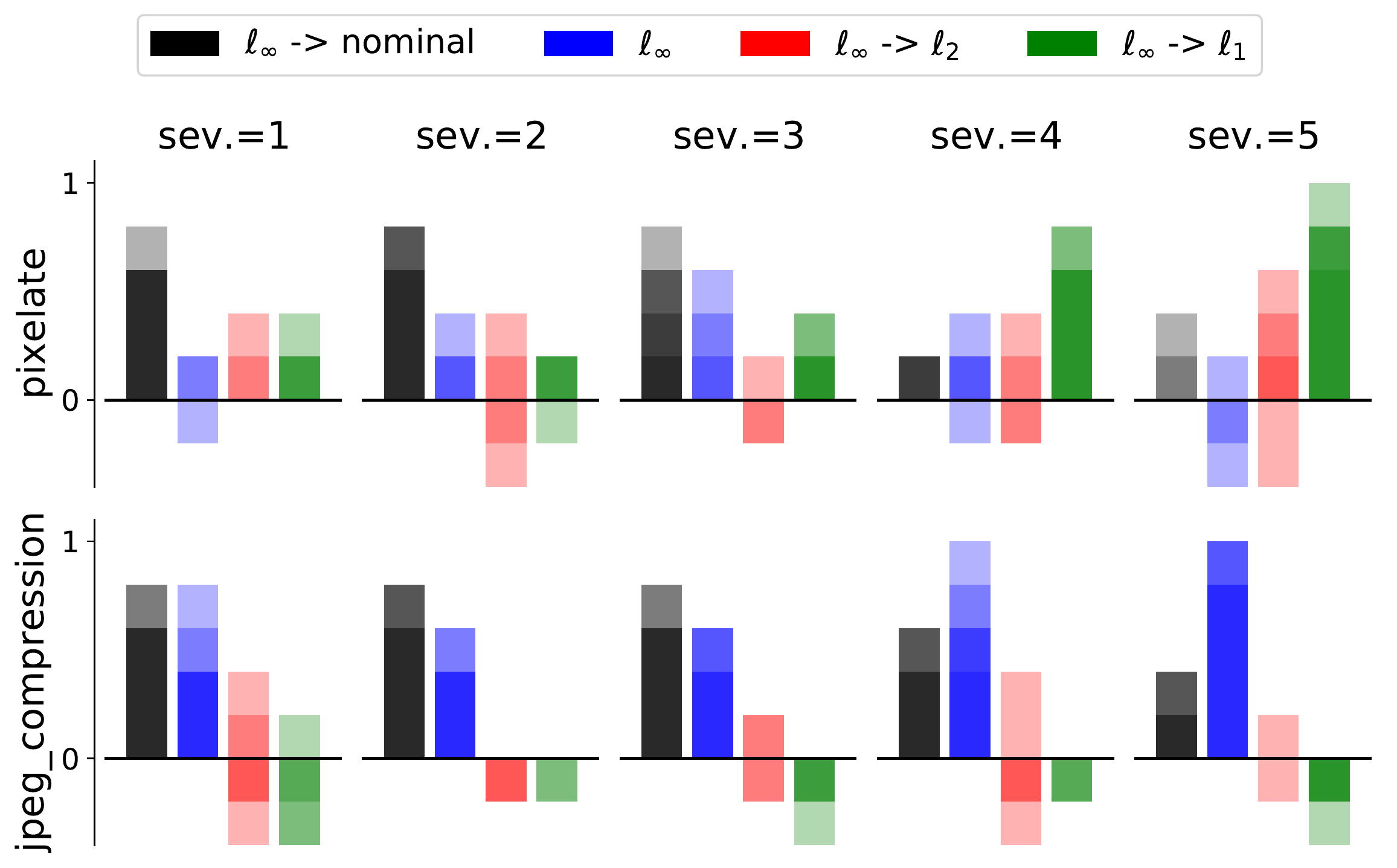}
\caption{
\textbf{Soup compositions on \imagenetc:} we plot the model-wise weights of the best soups across types and severities. 
} \label{fig:soup_composition_imagenetc}
\end{figure}

\subsection{A closer look at \imagenetc}
While our experiments have considered \imagenetc as a single dataset, it consists of 15 corruptions types, each with 5 severity levels.
As the various corruptions have different characteristics, one might expect the best soup to vary across them.
In Fig.~\ref{fig:soup_composition_imagenetc} we plot the composition of the top-5 soups for each severity level for two corruption types (as done in Fig.~\ref{fig:soups_composition}
).
The weights of the individual classifiers significantly change across distribution shifts: for both corruption types, increasing the severity (making perturbations stronger) leads to a reduction in the nominal weight in favor of a robust weight.
However, in the case of ``pixelate'' the soups concentrate on the $\ell_1$-robust network, while for ``jpeg compression'' this happens for $\ell_\infty$. Similar visualization for the remaining \imagenetc subsets are found in Fig.~\ref{fig:app_soup_composition_imagenetc_details} of the Appendix.
This highlights the importance of interpolating models with different types of robustness, and implies that considering each corruption type (including severity levels) as independent datasets could further improve the performance of the soups on \imagenetc.

\section{Discussion and Limitations}

Merging models with different types of robustness enables strong control of classifier performance by tuning only a few soup weights. 
Soups 
can find models which perform well even on distributions unseen during training (e.g. the \imagenet variants).
Moreover, our framework avoids co-training on multiple threats: this makes it possible to fine-tune models with additional attacks as they present themselves, and enrich the soups with them.

At the moment, our soups contain only nominal or $\ell_p$-robust models, but expanding the diversity of models might aid adaptation to new datasets.
We selected our soups with few-shot supervision, but other settings could potentially use soups, such as unsupervised domain adaptation~\cite{quinonero2009dataset,saenko2010adapting}, on labeled clean and unlabeled shifted data, and test-time adaptation~\cite{sun2020test,schneider2020improving,wang2021tent}, on unlabeled examples alone.
Moreover, in our evaluation we have constrained the soups to belong to a fixed grid, which might miss better models: future work could develop automatic schemes to optimize the soup weights, possibly with even fewer examples, or without labeled examples
(as done for test-time adaptation of non-robust models). 

\section{Conclusion}

We show that combining the parameters of robust classifiers, without additional training, achieves a smooth trade-off of robustness in different $\ell_p$-threat models.
This allows us to discover models which perform well on distribution shifts with only a limited number of examples of each shift.
In these ways, model soups serve as a good starting point to efficiently adapt classifiers to changes in data distributions.

{\small
\bibliographystyle{ieee_fullname}

}

\clearpage

\appendix

\section{Experimental details} \label{sec:app_exp_details}

\subsection{Training setup.}

\textbf{\cifar.} We train robust models from random initialization for 200 epochs with SGD with momentum as optimizer, an initial learning rate of 0.1 (reduced 10 times at epochs 100 and 150), and a batch size of 128.
For fine-tuning, we train for 10 epochs with cosine schedule for the learning rate, with peak value of 0.1 (we only use 0.5 for fine-tuning the model trained w.r.t. $\ell_1$ to the $\ell_2$-threat model) and linear ramp-up in the first 1/10 of training steps.
We generate adversarial perturbations by \apgd with 10 steps.
We select checkpoints according to robustness on a validation set as suggested by \cite{rice_overfitting_2020}.

\textbf{\imagenet.} 
We follow the setup of \cite{he_masked_2021}: for full training, we use 300 epochs, AdamW optimizer \cite{loshchilov2017decoupled} with momenta $\beta_1=0.9$, $\beta_2=0.95$, weight decay of 0.3 and a cosine learning rate decay with base learning rate $10^{-4}$ (scale as in \cite{goyal2017accurate}) and linear ramp-up of 20 epochs, batch size of 4096, label smoothing of 0.1, stochastic depth \cite{huang2016deep} with base value 0.1 and with a dropping probability linearly increasing with depth.
As data augmentation, we use random crops resized to 224 $\times$ 224 images, mixup \cite{zhang2017mixup}, CutMix \cite{yun2019cutmix} and RandAugment \cite{cubuk2019randaugment} with two layers, magnitude 9 and a random probability of 0.5.
We note that our implementation of RandAugment is based on the version in the \texttt{timm} library~\cite{rw2019timm}.
For \vit architectures, we adopt exponential moving average with momentum 0.9999.
For fine-tuning we keep the same hyperparameters except for reducing the base learning rate from $10^{-4}$ to $10^{-5}$ since this leads to better performance in the target threat model.
For adversarial training we use \apgd on the KL divergence loss with 2 steps for $\ell_\infty$- and $\ell_2$-norm bounded attacks, 20 steps for $\ell_1$ (as it is a more challenging threat model for optimization \cite{croce2022adversarial}).

\textbf{Baselines.}
For \Max and \sat, we fine-tune the singly-robust models with the same scheme above for the networks used in the model soups.
We generate adversarial perturbations with the the same attacks, and use 10 and 1 epoch of fine-tuning in the case of \cifar and \imagenet respectively.
For the baselines in Table~\ref{tab:imagenet_variants}, we use the same scheme (except for technique-specific components which follow the original papers).
For AdvProp we use dual normalization layers and train with random targeted attacks with bound $\epsilon=4/255$ on the $\ell_\infty$-norm.

\subsection{Evaluation setup.}

\begin{figure*}\centering\small
\begin{tabular}{c}
\rowcolor{Gray}

\rowcolor{Gray} 
soup: $\vtheta_{2 \rightarrow \infty} + \vtheta_{2} + \vtheta_{2 \rightarrow 1}$  
\\
\includegraphics[width=2\columnwidth]{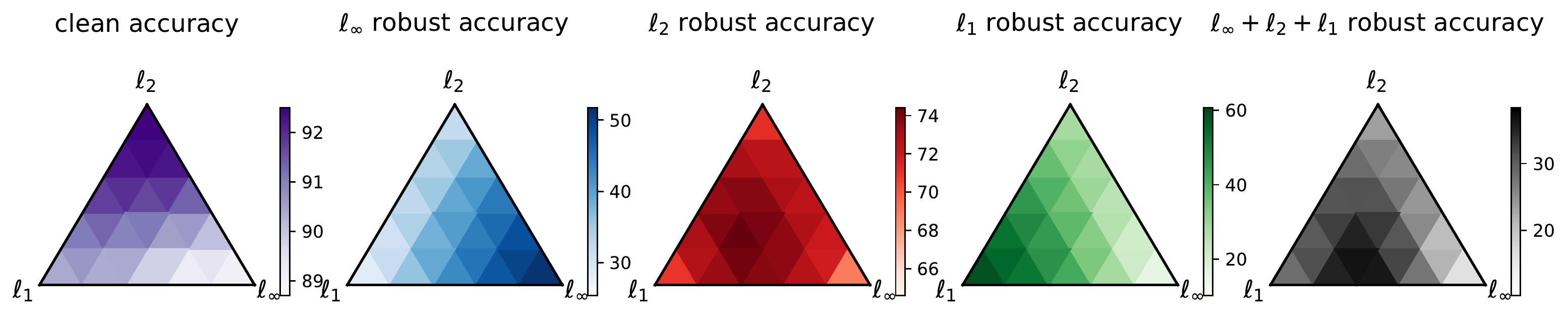}\\
\rowcolor{Gray} 
soup: $\vtheta_{1 \rightarrow \infty} + \vtheta_{1 \rightarrow 2} + \vtheta_{1}$
\\
\includegraphics[width=2\columnwidth]{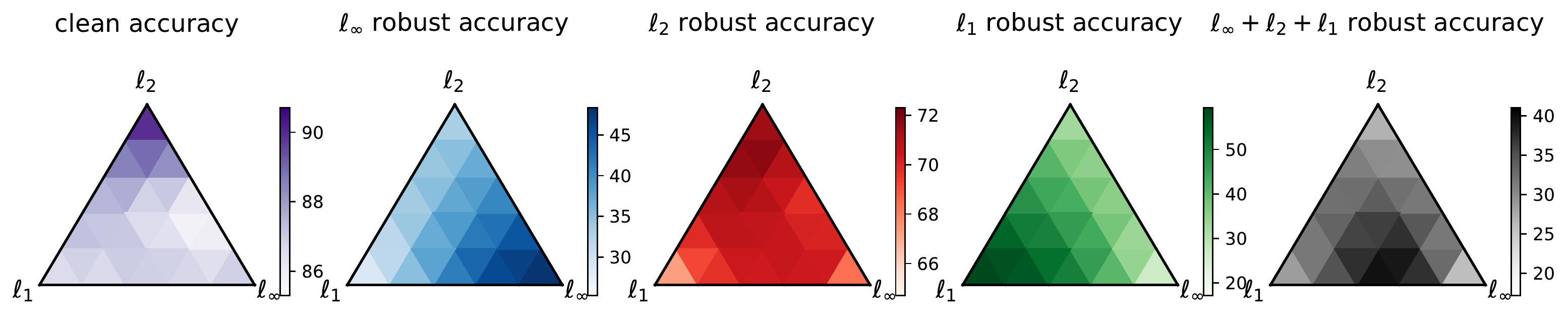}
\end{tabular}
\caption{
\textbf{Soups of three models on \cifar:} we fine-tune each model robust w.r.t. $\ell_p$ for $p \in \{2, 1\}$ (with \wrn-28-10 architecture) to the other threat models for 10 epochs, and show clean accuracy (first column) and robust accuracy w.r.t. every threat model (second to fourth columns) and their union (last column) of the soups obtained as convex combinations of the three bases.} \label{fig:app_soups_threemodels_cifar10}
\end{figure*}

\textbf{Adversarial robustness.} As default evaluation we use \apgd with 40 steps and default parameters, with the DLR loss \cite{croce_reliable_2020} for $\ell_\infty$- and $\ell_2$-attacks, cross entropy for $\ell_1$. As a test against stronger attacks, for the evaluation of the robustness of the soups of three threat models on \cifar (Fig.~\ref{fig:soups_threemodels_cifar10} and Fig.~\ref{fig:app_soups_threemodels_cifar10}) we increase the budget of \apgd to 100 steps and 5 random restarts (in this case we use targeted DLR loss for $\ell_\infty$ and $\ell_2$, and 1000 test points).

\textbf{Distribution shifts.} For all \imagenet variants we evaluate the classification accuracy on the entire dataset.

\section{Additional experiments}

\subsection{Soups with three threat models}
We show in Fig.~\ref{fig:app_soups_threemodels_cifar10} the clean accuracy, robust accuracy for each $\ell_p$-norm and their union of the soups obtained merging three classifiers. We use either a pre-trained classifier robust w.r.t. $\ell_2$ (top row) or $\ell_1$ (bottom row) and fine-tune them to the remaining threat models.

\begin{table*}
\centering \footnotesize \tabcolsep=2pt
\begin{tabular}{l|c|*{8}{C{13mm}}|C{13mm}}
    
    \cellcolor{header} \textsc{Setup} & \cellcolor{header} \textsc{\# FP} & \cellcolor{header} \textsc{ImageNet}  & \cellcolor{header} \textsc{IN-Real} & \cellcolor{header} \textsc{IN-V2} & \cellcolor{header} \textsc{IN-A} & \cellcolor{header} \textsc{IN-R} & \cellcolor{header} \textsc{IN-Sketch} & \cellcolor{header} \textsc{Conflict Stimuli} & \cellcolor{header} \textsc{IN-C} & \cellcolor{header} \textsc{Mean}   \\ \toprule

    \multicolumn{11}{l}{\textbf{Fixed grid search on 1000 images: best soups}} \\
    
    \toprule
    Single soup  & $\times 1$ & 82.49\% & 87.85\% & 71.99\% & 34.31\% & 53.84\% & 39.84\% & 38.52\% & 66.82\% & 59.46\%\\
    
    Dataset-specific soups & $\times 1$ & 82.29\% & 87.89\% & 71.95\% & 38.27\% & 56.39\% &40.73\% &67.03\% & 69.34\% & (64.24\%) \\
    \multicolumn{11}{l}{} \\
    
    \multicolumn{11}{l}{\textbf{Fixed grid search on 1000 images: second best soups}} \\ \toprule
    Single soup & $\times 1$ & 82.62\% & 87.92\% & 72.02\% & 30.99\% & 53.28\% & 39.00\% & 41.25\% & 68.75\% & 59.48\%\\
    
    Dataset-specific soups & $\times 1$ &82.49\%  & 87.84\%  & 71.94\% &  37.60\% & 56.42\% & 40.76\% & 66.95\% & 69.32\% & (64.16\%) \\
    \multicolumn{11}{l}{} \\
    
    \multicolumn{11}{l}{\textbf{Fixed grid search on 1000 images: third best soups}} \\ \toprule
    Single soup & $\times 1$ & 82.67\% & 87.86\% & 72.11\% & 34.25\% & 53.18\% & 39.52\% & 38.52\% & 67.60\% & 59.46\%\\
    
    Dataset-specific soups & $\times 1$ & 82.51\% & 87.65\% & 71.58\% &  37.51\% & 55.75\% & 40.65\% & 66.88\% &  68.92\% &(63.93\%) \\
    \bottomrule

    \end{tabular}
    
    \caption{\textbf{Top-k model soups for \imagenet variants:} we report the classification accuracy on the \imagenet variants, of the 1st, 2nd and 3d best soups (single or dataset-specific) found by grid search on the interpolation weights with 1000 points for each dataset.} \label{tab:app_imagenet_variants_topk_soups}
\end{table*}

\begin{table*}
\centering \footnotesize \tabcolsep=2pt
\begin{tabular}{l|c|*{8}{C{13mm}}|C{13mm}}
    
    \cellcolor{header} \textsc{Setup} & \cellcolor{header} \textsc{\# FP} & \cellcolor{header} \textsc{ImageNet}  & \cellcolor{header} \textsc{IN-Real} & \cellcolor{header} \textsc{IN-V2} & \cellcolor{header} \textsc{IN-A} & \cellcolor{header} \textsc{IN-R} & \cellcolor{header} \textsc{IN-Sketch} & \cellcolor{header} \textsc{Conflict Stimuli} & \cellcolor{header} \textsc{IN-C} & \cellcolor{header} \textsc{Mean}   \\ \toprule

    \multicolumn{11}{l}{\textbf{Robust models with standard $\epsilon_p$}} \\
    
    \toprule
    Single soup  & $\times 1$ & 82.49\% & 87.85\% & 71.99\% & 34.31\% & 53.84\% & 39.84\% & 38.52\% & 66.82\% & 59.46\%\\
    
    Dataset-specific soups & $\times 1$ & 82.29\% & 87.89\% & 71.95\% & 38.27\% & 56.39\% &40.73\% &67.03\% & 69.34\% & (64.24\%) \\
    \multicolumn{11}{l}{} \\

    \multicolumn{11}{l}{\textbf{Robust models with $\epsilon_p / 2$}} \\ \toprule
    
    Single soup & $\times 1$& 82.66\% & 87.78\% & 72.34\% & 32.05\% & 51.40\% & 37.80\% & 39.69\% & 69.06\% & 59.10\% \\
    
    Dataset-specific soups & $\times 1$ &81.85\% & 87.47\%&72.34\% &36.25\% &53.80\% &39.32\% &62.19\% & 69.44\% &(62.83\%) \\
    \multicolumn{11}{l}{} \\ 
    \multicolumn{11}{l}{\textbf{Robust models with $\epsilon_p / 4$}} \\ \toprule
    
    Single soup  & $\times 1$& 81.47\% & 87.36\% & 70.84\% & 26.23\% & 53.46\% & 39.13\% & 46.25\% & 67.37\% & 59.01\%\\
    
    Dataset-specific soups & $\times 1$ & 82.68\% & 87.72\% &72.21\% & 35.61\% &54.00\% &39.79\% & 59.22\%& 69.49\%& (62.59\%) \\
    \bottomrule
    \end{tabular}
    
    \caption{\textbf{Varying threat models:} we report the classification accuracy on the \imagenet variants 
    of the best single and dataset-specific soups when using various radii $\epsilon_p$ for fine-tuning the $\ell_p$-robust networks.
    The soups are selected via a fixed grid search on the interpolation weights with 1000 points for each dataset.} \label{tab:app_imagenet_variants_varying_eps}
\end{table*}

\subsection{Model soups on \imagenet variants}

We show additional results for model soups on \imagenet variants: first, in Table~\ref{tab:app_imagenet_variants_topk_soups} we report the results on the full datasets of the second and third best soups according to the grid search on 1000 points for the shift (we also show the best soup from Table~\ref{tab:imagenet_variants} in the main part). When selecting the best soup on average across datasets, all three classifiers have very close performance (59.46\% to 59.48\%), while the accuracy on the individual datasets may vary e.g. on \imageneta and \conflictstimuli. For the dataset-specific soups the results on the entire datasets respect the ranking given by the grid search (the three values are similar to each other), further suggesting that even a limited number of points can serve to tune a suitable soup.

Second, we study the effect of varying the radii of the $\ell_p$-threat models used by the robust classifiers in the soups. In this case, we fine-tune for 1 epoch the original model robust w.r.t. $\ell_\infty$ at $\epsilon=4/255$ with adversarial training w.r.t. $\ell_\infty$ at $\epsilon_\infty \in \{1/255, 2/255\}$, $\ell_2$ at $\epsilon_2 \in \{1, 2\}$ ($\epsilon_2=4$ above), $\ell_2$ at $\epsilon_1 \in \{64, 128\}$ ($\epsilon_1=255$ above). In this way we have three sets of four models to create soups, where the nominal one is fixed and the robust ones have radii $\epsilon_p$, $\epsilon_p / 2$ and $\epsilon_p / 4$ for $p\in\{\infty, 2, 1\}$. Table~\ref{tab:app_imagenet_variants_varying_eps} reports the results of the various sets of models: for the single soup optimized for average performance, the smaller $\epsilon_p$ slightly reduce the performance. Looking at the individual datasets, in some cases like \imagenetvtwo and \imagenetc using smaller values of $\epsilon_p$ yields some improvements, but it also leads to severe drops on the distribution shifts where having robust models is more relevant like \conflictstimuli and \imagenetr. This suggests that it might be useful to have models robust w.r.t. the same $\ell_p$-norm but with different radii in the set of the networks used for creating the soups.

\begin{figure*}\centering\small
\begin{tabular}{cc}
\includegraphics[align=t, width=1\columnwidth]{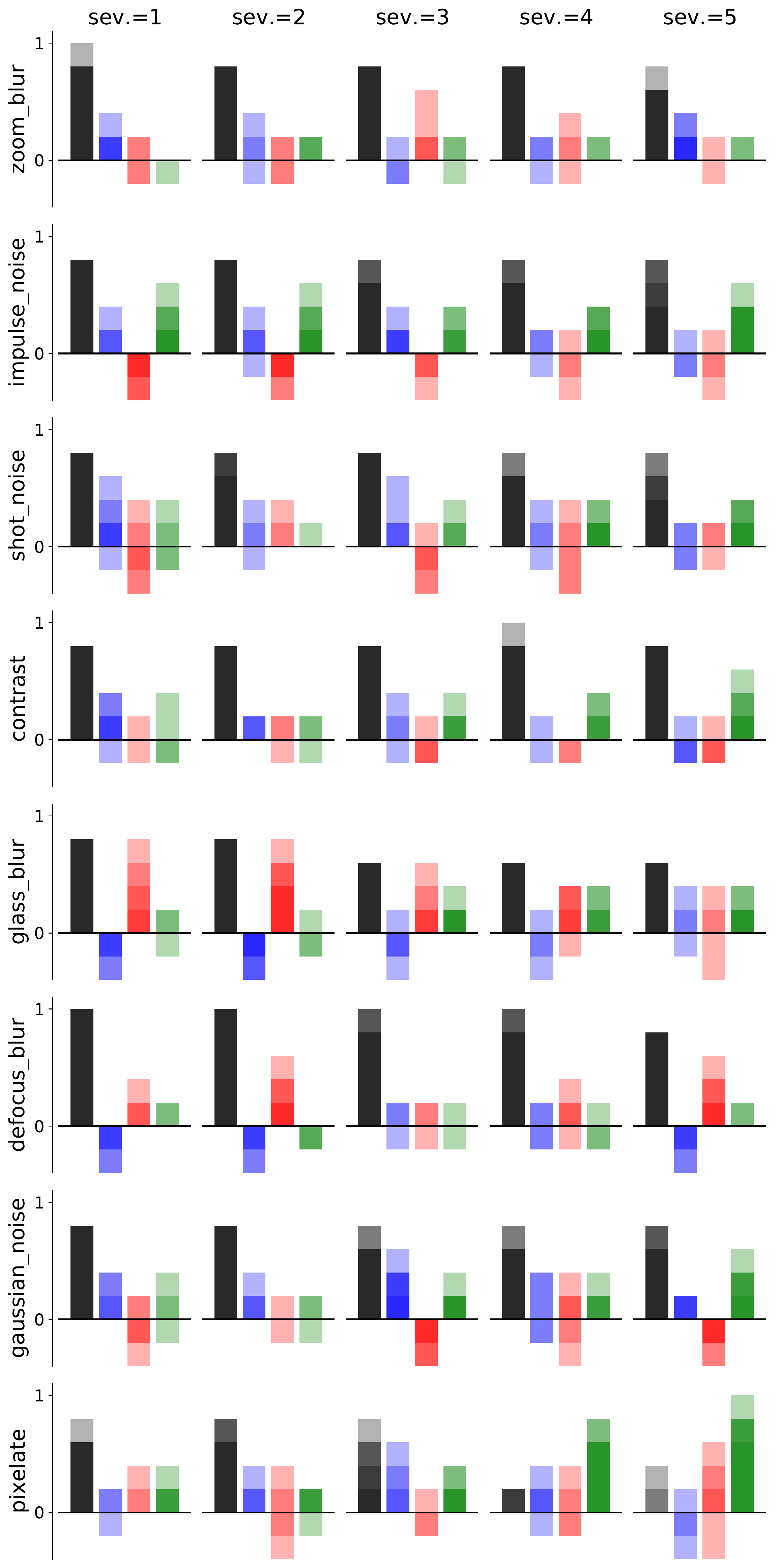} & \includegraphics[align=t, width=1\columnwidth]{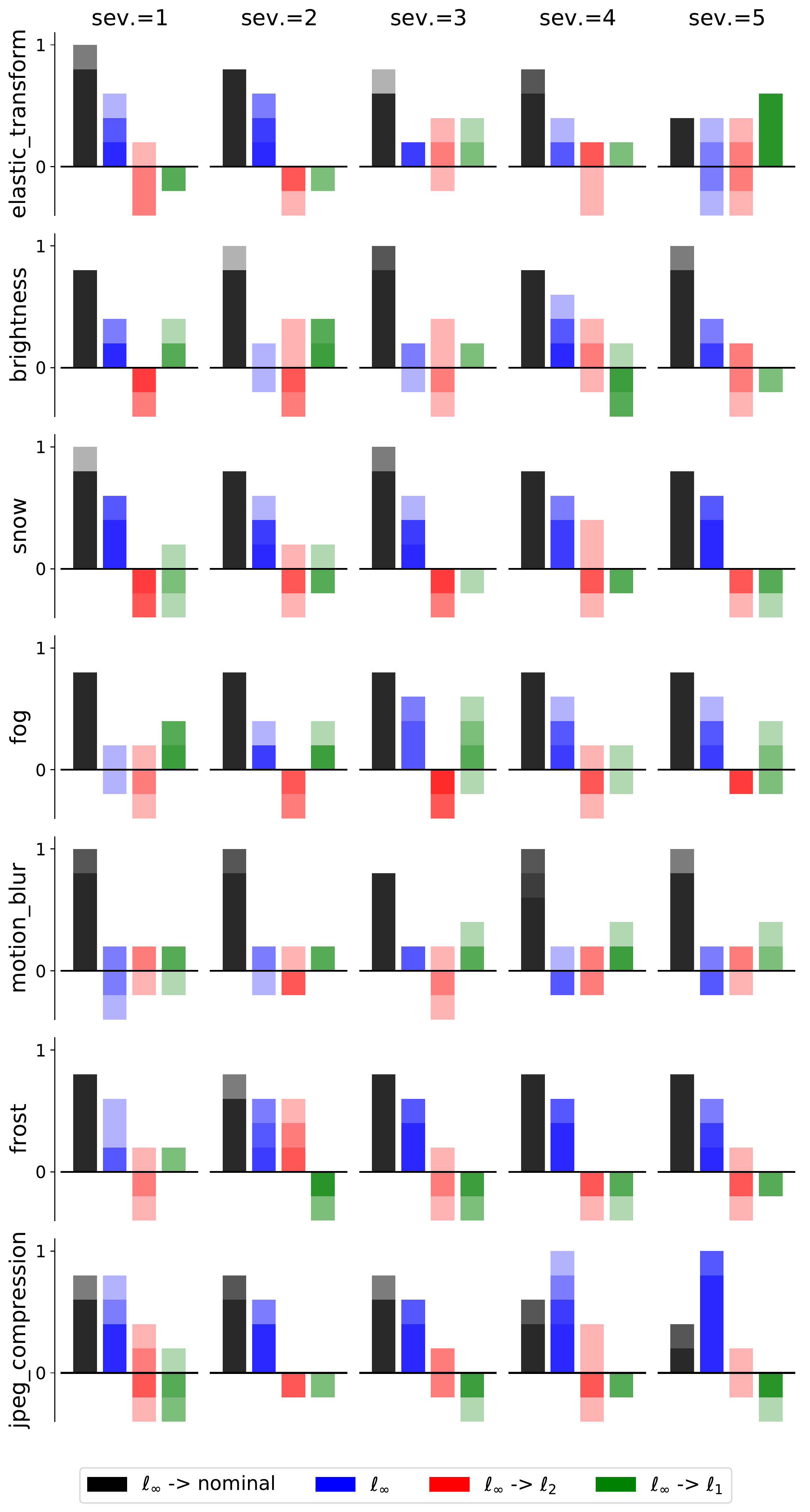}
\end{tabular}
\caption{\textbf{Best soups over \imagenetc subsets:} we plot the composition of the top 5 soups found by the grid search for each corruption type and severity level.} \label{fig:app_soup_composition_imagenetc_details}
\end{figure*}

\subsection{Composition of soups on \imagenetc}
In Fig.~\ref{fig:app_soup_composition_imagenetc_details} we visualize the composition of the top-5 soups for each corruption type and severity level: one can observe that the weights of the four networks in the soups varies across \imagenet subset.

\end{document}